\documentclass[final,3p,times,twocolumn]{elsarticle}

\usepackage{amssymb}
\usepackage{amsmath}
\usepackage{algpseudocode}
\usepackage{svg}
\usepackage{algorithm}
\usepackage{lineno}
\usepackage{bm} 
\usepackage{longtable}
\usepackage{caption}
\usepackage{algorithmicx}

\usepackage{adjustbox}
\usepackage{longtable}
\usepackage{subfig}
\usepackage{soul}
\usepackage{makecell}
\usepackage{color}
\usepackage{xtab}
\usepackage{booktabs}
\usepackage{algorithm}
\usepackage{algpseudocode}
\usepackage{tabularx}

\usepackage{array}
\newcolumntype{L}{>{\raggedright\arraybackslash}c}

\usepackage[normalem]{ulem}

\usepackage{lineno}

\journal{Annals of Nuclear Energy}

\begin{document}

\begin{frontmatter}



\title{Gradient-based optimization of nuclear criticality experiments\\using neural surrogate eigenvalue sensitivities}


\author{Will Savage}
\author{Logan Burnett}
\author{Dean Price*}
\ead{dnprice@mit.edu}

\affiliation{organization={Department of Nuclear Science and Engineering, Massachusetts Institute of Technology},
            addressline={60 Vassar St.}, 
            city={Cambridge},
            postcode={02139}, 
            state={MA},
            country={US}}

\begin{abstract}

The validation of advanced nuclear reactor designs and fuel concepts will require the design of new critical experiments with high neutronic similarity to the target technology.
Neutronic similarity can be quantified by the correlation coefficient $c_k$, which captures the shared bias in $k_\text{eff}$ induced by uncertainties in nuclear data.
Generally, a $c_k\geq0.9$ is needed for an experiment to be sufficiently similar to a target technology.
In this work, a physics-informed deep neural network is trained to predict the neutronic sensitivity of grid-based critical experiment geometries.
The differentiability of the neural network is used to enable gradient-based design optimization of new experiment geometries to maximize $c_k$ with the sensitivity profile of a target technology.
This approach allows for optimization over the combinatorial design space of potential material combinations within the grid, moving beyond traditional parametric optimization approaches. 
As a proof-of-concept optimization exercise, the method is applied to the validation of the TN-Americas TN-LC transportation cask with HALEU fuel, for which existing critical experiment coverage is limited.
This application is shown to produce experiment geometries achieving $c_k$ scores of 0.97757, 0.81324, and 0.93276 for three configurations of interest.

\end{abstract}




\end{frontmatter}



\section{Introduction}
\label{sec:intro}
Driven by the rapid scaling of AI data centers, domestic reindustrialization, and a growing desire for energy independence, the public and private sectors alike are turning to nuclear power to meet the energy demands of the coming century~\cite{intro1}~\cite{intro2}~\cite{intro3}. 
The revitalization of the nuclear industry has led to the introduction of a plethora of advanced reactor designs, fuels, and materials.
Critical experiments play a central role in supporting the validation and eventual deployment of these technologies.

Critical (or criticality) experiments are low-power assemblies of fissile material designed to precisely measure and validate the neutronic properties of a nuclear system without requiring an expensive and difficult full-scale demonstration~\cite{thompson2023}. 
The National Criticality Experiment Research Center (NCERC) supports criticality experiments for a variety of applications with multiple critical assemblies. 
The Comet~\cite{thompson2021} and Planet~\cite{sanchez2021} vertical-lift assemblies at NCERC have been used to conduct a variety of criticality experiments. 
Both assemblies are general-purpose, heavy-duty, vertical-lift assemblies made of two vertically stacked platforms, each containing a subcritical fissile configuration. 
Reactivity is controlled by changing the distance between the platforms.
Historically, the Zero Power Reactor (ZPR) and Zero Power Plutonium/Physics Reactor (ZPPR) facilities were a series of Horizontal Split-Table (HST) critical experiment facilities operated at Argonne National Laboratory between 1955 and 1990~\cite{zpr}.
The ZPR/ZPPR approach to critical experiment design allowed for a range of experiments to be conducted within a single assembly by using a grid of interchangeable material tubes.
Interest in advanced reactor designs has demonstrated the need for a new critical experiment machine capable of supporting advanced fuels, geometries, and material compositions. 
The System Physics Advanced Reactor Critical facility (SPARC) project at Idaho National Laboratory is a proposed critical experiment facility that will support this need. 
Unlike the vertical-lift mechanisms of Comet and Planet, SPARC will be an HST machine capable of supporting a critical assembly with a mass of 24,000 kg and a length of around 2.0 m in each dimension. 
SPARC will be crucial to the validation and deployment of advanced reactor designs and fuel concepts~\cite{woolstenhulme2025}.

To yield useful results, the neutronic properties of a criticality experiment must be similar to those of the full-scale system. 
Specifically, the uncertainties associated with each cross section involved in the full-scale system should be targeted by the criticality experiment to encourage any calculation biases to manifest similarly in both systems.
This similarity can be measured by the correlation coefficient $c_k$ described in Section \ref{sec:method}.
Other proposed quantifiers of neutronic correlation include the Accumulative Correlation Coefficient for Relevance of Uncertainties in Experimental (ACCRUE) ~\cite{accrue} similarity coefficient, introduced by Seo et al. in 2021.
An advantage of ACCRUE is that it accounts for experimental measurement uncertainties, which $c_k$ does not.
Additionally, $c_k$ has been criticized due to its sensitivity to evaluation-dependent and sometimes unreliable covariance data.
However, it is used in this study due to its comparative historical prevalence in the field, as well as its prior application to validation of HALEU transportation packages~\cite{eidelpes2019}.

If a criticality experiment and full-scale system have a high $c_k$, observed neutronic properties of the experiment can be safely transferred to the full-scale system.
In the context of experiment design, Maldonado et al.~\cite{maldonado2023} rely on $c_k$ values to guide the design of criticality experiments for the conceptual Snowflake microreactor. 
As the correlation coefficient between an experiment design and Snowflake approaches unity, biases in calculated $k_\text{eff}$ due to cross-sectional uncertainty associated with the experiment approach those associated with the commercial technology. 
This work demonstrates how $c_k$ can be used as an objective similarity metric for experiment design.
Additionally, Walters \& Roskoff~\cite{walters2026designing} used both sensitivities of $k_{\text{eff}}$ and reactivity to cross sections in order to evaluate the similarity between a 30 MWt graphite-moderated test reactor and a criticality experiment.

Many advanced reactor designs require high-assay low-enriched uranium (HALEU) as fuel, which will require safe transportation solutions. 
The TN Americas TN-LC cask~\cite{eidelpes2019} is a representative transportation package that could be plausibly adapted to support HALEU, avoiding the need to initiate a new design and regulatory approval process. 
To validate TN-LC for HALEU transportation, rigorous neutronic analysis must be carried out to ensure subcriticality in all conditions including situations such as water leakage. 
To conduct neutronic analysis of the TN-LC, Eidelpes et al.~\cite{eidelpes2019} use correlation coefficients to identify benchmark critical experiments with high $c_k$. 
Hall et al.~\cite{hall2020} consider two systems to be similar if they have a $c_k \geq 0.9$, while Eidelpes et al. require that a $k_\text{eff}$ penalty be applied to subcritical margins without $c_k >0.8$. 
130 experiments were found with $c_k \geq 0.8$ for Case 1, but none with $c_k \geq 0.9$, and none with $c_k \geq 0.8$ for Case 2.
This demonstrates the need for new experiments to be designed. 

Machine learning techniques have been explored to optimize criticality experiment design. 
One such technique, Constrained Bayesian Optimization (CBO), strategically selects experimental configurations to evaluate while maintaining a probabilistic model of the objective function, minimizing expensive evaluations while enforcing the physical constraints of the design space. 
Branco-Ketcher et al.~\cite{branco2025} apply CBO to the design of criticality experiments for TerraPower's Molten Chloride Reactor Experiment (MCRE) to simultaneously maximize $c_k$ and sensitivity to reactions of interest while remaining within the physical limits of the Comet assembly.
Targeting the reduction of intermediate-energy Pu-239 nuclear data, Kleedtke et al. \cite{kleedtke2025} used particle swarm optimization to adjust an experimental configuration such that the neutron flux and sensitivity of $k_\text{eff}$ to the total fission cross section are both maximized in the 1 keV to 600 keV energy range.
Michaud et al.~\cite{gp} use a Gassian Process (GP) optimization, which explores the design space of critical experiments with the advantage of providing uncertainties associated with each design.
Given the extreme potential computational cost of Monte Carlo based methods for evaluating each candidate model during the optimization procedure, a simplified model was used with limited sampling parameters.
An alternative approach to address this computational cost concern is to use a surrogate model capable of rapidly approximating the objective function.

Surrogate models to replace the high computational cost of full-order model evaluations within optimizers have seen frequent use in the nuclear engineering field.
Whyte and Parks \cite{whyte2021} evaluated two surrogate models to be used in optimizing the design of a 36 assembly layout in a miniaturized pressurized water reactor.
These two deep learning based surrogate models were then embedded in a NSGA-II \cite{deb2002fast} multiobjective optimizer.
Separately, a two part study \cite{seurin2025techno, seurin2026techno} optimized levelized cost of electricity for a heat pipe microreactor while maintaining constraints on safety-relevant quantities.
Advanced reinforcement learning-based multiobjective optimization methods were then applied to treat these constraints in a more comprehensive manner. 
In both cases, neural networks were used as surrogate models.
One important practice used was to reintroduce the full-order model at the conclusion of the optimization routine to evaluate the accuracy of the surrogate models specifically on the discovered optimal solution.
In general, this practice is fairly common \cite{price2022multiobjective, sabater2020efficient, nikolaou2025multi} and acts to mitigate some of the drawbacks arising from inaccuracies of any surrogate model.

In reference to the methods behind the surrogate models themselves, neutron transport calculations can be nonlinear and therefore difficult to accurately model with simple surrogate methods. 
However, advances in deep learning~\cite{lecun2015} over recent years have produced architectures capable of approximating complex functions ranging from protein folding \cite{xu2019distance} to fluid dynamics \cite{kutz2017deep}, making deep neural networks a compelling candidate for neutronic surrogate modeling. 
A number of deep learning techniques can be applied to build strong neutronic surrogate models.
Convolutional layers, for which Convolutional Neural Networks (CNNs) are named, are highly adept at extracting spatial and geometric information from multidimensional inputs, which has led to their ubiquitous presence in image-processing models over the past few decades~\cite{lecun1998}. 
Attention, the primary architectural advancement behind the emergence of Large Language Models, learns the relationships between all components of an input simultaneously to capture long-ranged effects ~\cite{vaswani2017}. 
When combined in a single neural network, these components present a powerful and promising method for capturing the neutronic behavior of complex geometries in a fraction of the time of a full Monte Carlo transport simulation.

Deep neural networks are differentiable objects, meaning that they can be used with gradient-based optimization methods.
Gradient-based optimization methods have empirically demonstrated strong optimization capabilities throughout machine learning, and scale well with the dimensionality of the optimization space~\cite{williams2023}.
Backpropagation~ can be used to compute the gradient of the objective with respect to each component of the input.
The gradient is then used to update the input in the direction that maximally increases or decreases (depending on the formulation of the problem) the objective. 
In the context of nuclear system design, Pevey et al.~\cite{pevey2022} use gradient methods to optimize the neutronic properties of a simplified core geometry. 
If a differentiable surrogate for neutronic sensitivity profiles can be constructed, gradient-based optimization can be applied to critical experiment design.  

This work exploits the differentiability of deep neural network surrogates to perform gradient-based optimization of $c_k$ over the full combinatorial space of material assignments in a voxelized experimental geometry. A model of the target technology is first built in OpenMC and used to calculate target sensitivity profiles for various scenarios of interest. A dataset of procedurally generated material configurations is then used to train a surrogate neural network to predict sensitivities for arbitrary geometries. The surrogate is comprised of a novel, heterogeneous neural network architecture designed to promote an inductive bias of the underlying transport physics during training. Once trained, the surrogate is used with a gradient-based optimization method to generate experiment geometries that maximize $c_k$ for any input sensitivity profile generated by OpenMC.

\begingroup{}
This work is a proof-of-concept optimization exercise that evaluates the preliminary performance of an unconstrained optimization method for maximizing $c_k$.
The optimization is carried out within a fixed design space, where the geometry size, material palette, symmetry, and grid discretization are held constant.
The goal of this work is to determine whether a differentiable surrogate can guide an optimizer towards geometries with high $c_k$ in a controlled setting.
\endgroup{}

\section{Methodology}
\label{sec:method}
\subsection{Application Scenario}
\label{subsec:app}
The TN Americas TN-LC transportation cask~\cite{eidelpes2019} was originally developed for transporting moderate to high-burnup spent nuclear fuel (SNF) assemblies, and is now being considered for future transportation of HALEU needed by advanced reactors. 
If selected for this purpose, criticality safety evaluations would be required to show that it remains subcritical in all possible scenarios, including events of water intrusion into the cask and/or internal $\text{UO}_2$ canisters. 
Data from criticality experiments with high neutronic similarity to the TN-LC can be used to aid this analysis.
The neutronic similarity between an experiment and the target system can be quantified by the correlation coefficient $c_k$, which is the fraction of nuclear data-induced uncertainty that the systems share~\cite{maldonado2023}. 
A sensitivity profile is constructed by combining the sensitivity vectors $s \in \mathbb{R}^G$ for the reactions of interest into a single vector $S \in \mathbb{R}^{RG}$ for $R$ reactions and $G$ discrete energy groups. The correlation coefficient is defined as 
\begin{equation}
\label{eq:ckdefinition}
    c_k
    =
    \frac{
    S_{\mathrm{TARGET}}^\top C_{XX}S
    }{
    \sqrt{
    \left(S_{\mathrm{TARGET}}^\top C_{XX}S_{\mathrm{TARGET}}\right)
    \left(S^\top C_{XX}S\right)
    }
    }
\end{equation}
where $S_{TARGET}$ is the sensitivity profile of the target, $S$ is the sensitivity profile of the experiment, and $C_{XX}$ is a matrix containing the relative covariances of the nuclear data.

In their analysis of a different candidate transportation cask, Hall et al.~\cite{hall2020} consider two systems with a $c_k \geq 0.9$ to be ``neutronically similar.'' 
In their evaluation of the TN-LC, Eidelpes et al.~\cite{eidelpes2019} use the TSUNAMI code to find existing critical experiment benchmarks with high $c_k$ for two limiting configurations of the cask.
In both cases, the fuel canisters are flooded while the cask cavity remains dry. 
In Case 1, the cask is modeled with a low-density foam in the cask cavity and low $^{10}\text{B}$ areal density in the canister poison coating. 
In Case 2, the cask is modeled with a medium density foam and a high $^{10}\text{B}$ areal density.
While TSUNAMI identified 130 critical experiments with $c_k \geq 0.8$ for Case 1, it identified no experiments with $c_k \geq 0.8$ for Case 2, and none with $c_k \geq 0.9$ for either Case 1 or Case 2.
The lack of highly correlated experiments for the TN-LC highlights the need for new critical experiments to be designed, especially given the capability of the SPARC facility to accommodate experiments involving HALEU.

\begin{table}[H]
\centering
\renewcommand{\arraystretch}{1.2}
\begin{tabular}{@{}lp{0.65\columnwidth}@{}}
\hline
\textbf{Configuration} & \textbf{Description} \\
\hline
Dry Cask & No water intrusion; no foam in cavity; low $^{10}$B areal density. \\
Case 1   & Flooded canisters; low-density foam in cavity; low $^{10}$B areal density. \\
Case 2   & Flooded canisters; medium-density foam in cavity; high $^{10}$B areal density. \\
\hline
\end{tabular}
\caption{Brief descriptions of the three TN-LC configurations considered in this work.}
\label{tab:config_summary}
\end{table}
\begin{figure}[H]
    \centering
    \includegraphics[width=1\linewidth]{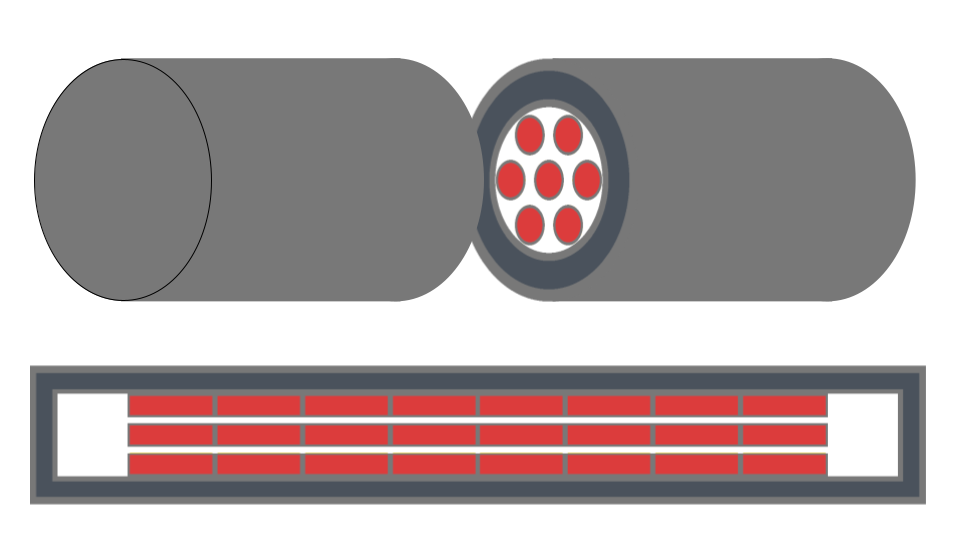}
    \caption{Renderings of the simplified TN-LC model in OpenMC showing the arrangement of HALEU canisters (red) within the cask.}
    \label{fig:model}
\end{figure}
A model of the TN-LC was built in both OpenMC and Serpent~\cite{serpent} following a simplified design presented in Eidelpes et al. and used to calculate sensitivity profiles for Case 1, Case 2, and a ``dry'' configuration of the cask without any water intrusion. 
Table \ref{tab:config_summary} is provided as a concise summary of these cases.

Serpent does not support reflective cylindrical boundaries, so a cuboid reflective boundary was placed around each cask model, unlike the cylindrical reflective boundary used by Eidelpes et al~\cite{eidelpes2019}.
As discussed in Section~\ref{subsec:train}, model agreement between the codes was necessary to benchmark the sensitivity calculation implementation in OpenMC against that of Serpent, which required the model to adhere to the limitations of the Serpent boundary conditions.

The simplified model consists of an outer cylinder of 304 stainless steel, a layer of lead inside the outer steel cylinder, and an inner steel cylinder inside the lead layer.
There are 56 steel fuel canisters within the inner cylinder, each coated in a thin layer of $\text{AlB}$ neutron poison and containing 20\% enriched $\text{UO}_2$.
The external neutron shield resin, aluminum shield boxes, and impact limiters are not modeled.
Cross-sectional cutaways of the model are shown in Figure~\ref{fig:model}.

\begin{figure*}[tb]
    \centering
    \includegraphics[width=\textwidth]{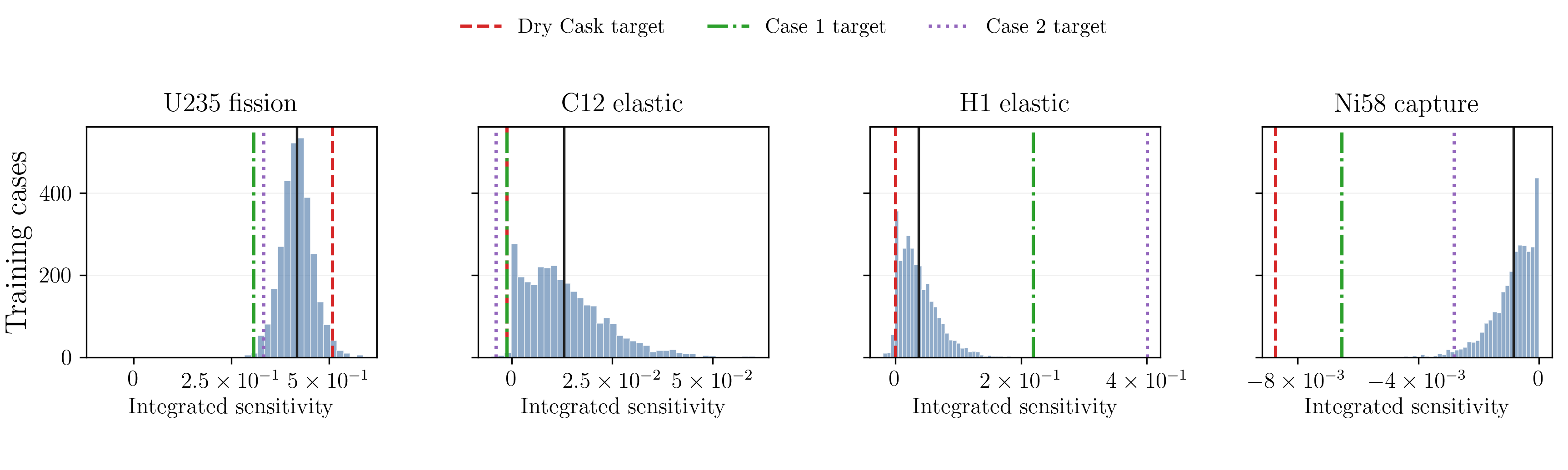}
    \caption{Integrated $k_\text{eff}$ sensitivities of the Dry Cask, Case 1, and Case 2 TN-LC targets, shown as vertical lines on the distributions and means of the 3,000 training geometries.}
    \label{fig:int_sens_targets}
\end{figure*}

\begin{figure}[H]
    \centering
    \includegraphics[width=1\linewidth]{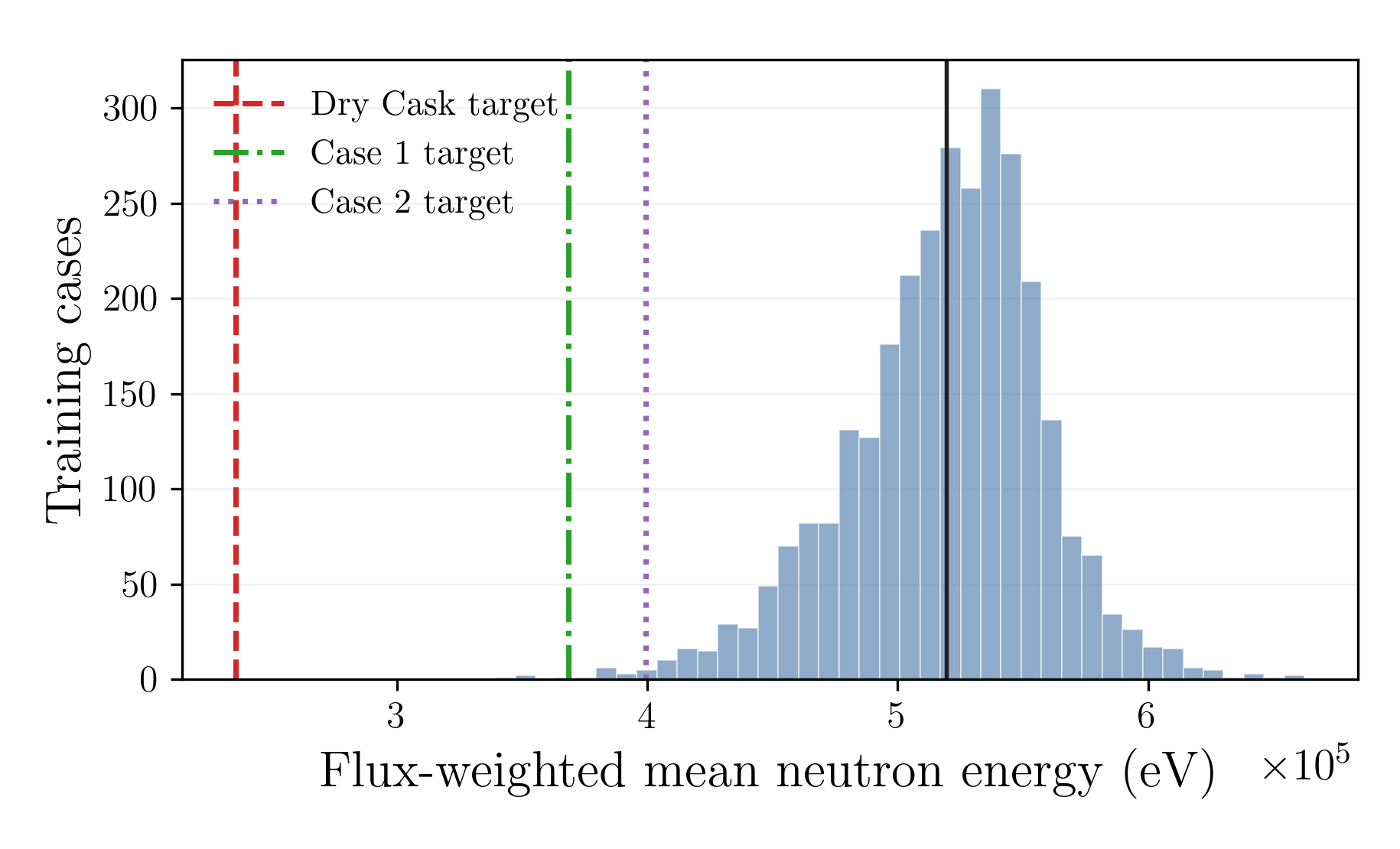}
    \caption{Flux-weighted mean neutron energy $\bar E$ of the Dry Cask, Case 1, and Case 2 TN-LC targets, shown as vertical lines on the distribution and mean of the 3,000 training geometries.}
    \label{fig:flux_targets}
\end{figure}

\begin{figure*}[tb]
    \centering
    \includegraphics[width=0.8\textwidth]{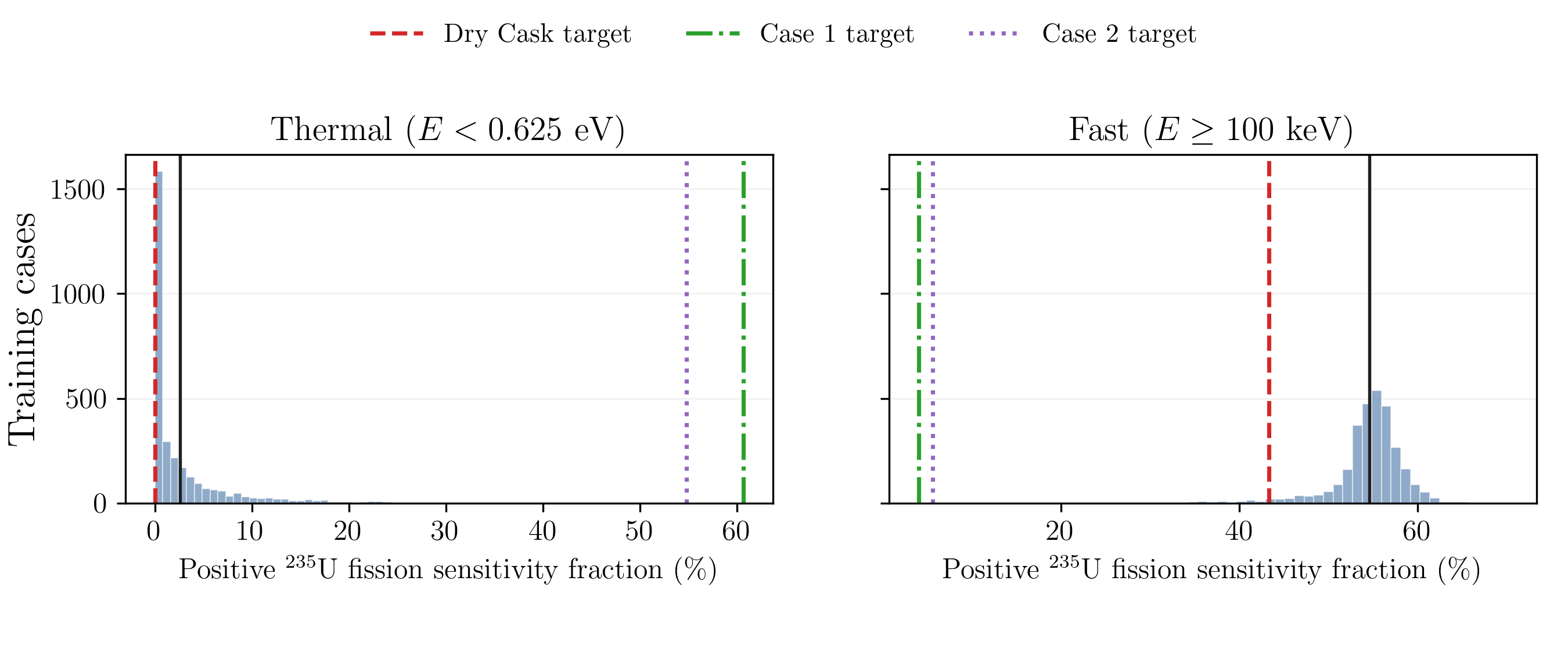}
    \caption{Thermal and fast fractions of the positive $^{235}$U fission sensitivity for the Dry Cask, Case 1, and Case 2 TN-LC targets, shown as vertical lines on the distributions and means of the 3,000 training geometries.}
    \label{fig:spec_targets}
\end{figure*}

\begin{figure}[H]
    \centering
    \includegraphics[width=0.7\linewidth]{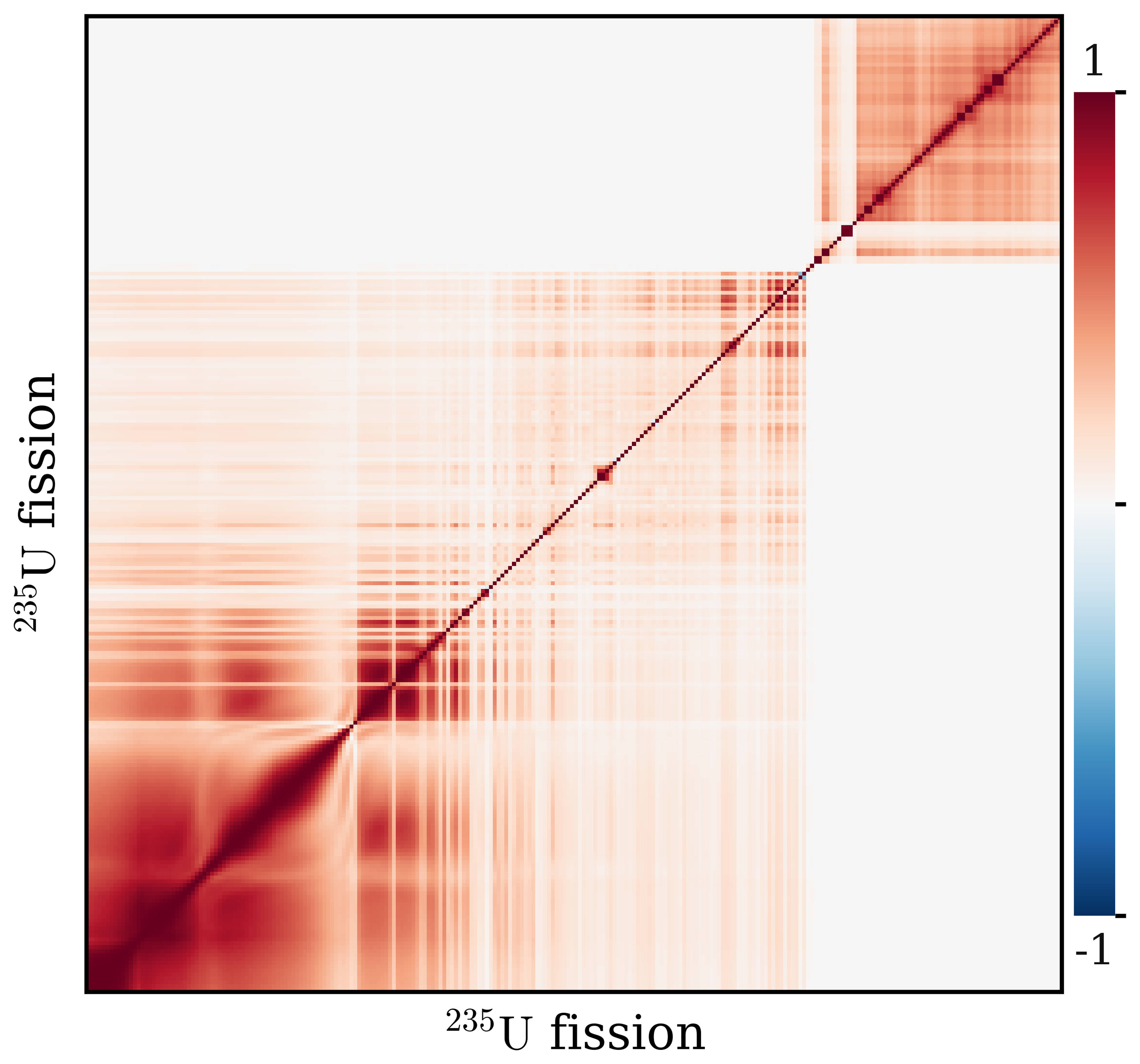}
    \caption{Correlation matrix for uncertainties in the $^{235}\text{U}$ fission cross-section derived from the ENDF/B-VII.1 nuclear data library using the SCALE-252 energy group structure.}
    \label{fig:cor}
\end{figure}

Eidelpes et al. present numerous dry cask configurations with different neutron poison arrangements.
The configuration modeled in this work had the highest $k_\text{eff}$ of all the options, and was thus the most relevant from a criticality safety perspective.
In Case 1 and Case 2, the void within the cask is filled with polyurethane foam, which is not present in the dry cask model.
Cask water intrusion is modeled via a homogeneous mixture of $\text{UO}_2$ and $\text{H}_2\text{O}$ determined by the desired volume ratio of each case.
The reactions chosen for the sensitivity profiles were $^{235}\text{U}$ fission, $^{12}\text{C}$ elastic scattering, $^{1}\text{H}$ elastic scattering, and $^{58}\text{Ni}$ absorption. 
As shown in Section~\ref{sec:results}, sensitivity to the $^{235}\text{U}$ fission cross-section is the primary contributor to $k_\text{eff}$ uncertainty.
The $^{1}\text{H}$, $^{12}\text{C}$, and $^{58}\text{Ni}$ sensitivities are included as representative sensitivities for the other materials present in the model.
Modeling details and sensitivities for the different TN-LC configurations are shown in Section~\ref{subsec:cask}.
\begingroup{}
The neutronic properties of the three target configurations are compared to those of the surrogate training data in Figures~\ref{fig:int_sens_targets},~\ref{fig:flux_targets}, and~\ref{fig:spec_targets}, using the descriptors defined in Section~\ref{subsec:train}.
The thermal ($E<0.625\,$eV) and fast ($E\geq100\,$keV) fractions of the positive $^{235}$U fission sensitivity are also shown in Figure~\ref{fig:spec_targets} to describe the energy distribution of fission importance.
The flooded Case 1 and Case 2 targets are strongly thermal, with more than half of their positive $^{235}$U fission sensitivity in the thermal range, far exceeding that of any training geometry.
The flux-weighted mean energy of the Dry Cask target is also below that of any training geometry, while those of Case 1 and Case 2 lie in the low-energy tail of the training distribution.
Additionally, the integrated $^{1}$H elastic sensitivity of Case 2 and the integrated $^{58}$Ni capture sensitivities of the Dry Cask and Case 1 lie outside the range of the training data.
These comparisons identify regions of the parameter space that are relevant to the application cases, but are not covered by the training data.
\endgroup{}
To enable proper calculation of covariance-weighted $c_k$ scores, the uncertainty covariances and correlations for the four reactions were extracted from the SCALE-6.3.2~\cite{scale} code using the CADILLAC and COGNAC AMPX utility modules.
The correlation matrix for uncertainties in the $^{235}$U fission cross section is shown in Figure~\ref{fig:cor}, using the SCALE-252 energy group structure.
The fine-energy SCALE-252 structure is used throughout this study due to its complex covariance data as well as the handling of the complex, high-uncertainty resonance structures in reactions such as $^{235}$U fission, which present an interesting target for the optimization method.
The ENDF/B-VIII.0~\cite{endf8} Evaluated Nuclear Data Library is used in this work for all transport calculations, but covariances are drawn from the SCALE6.3.2 covariance library based on ENDF/B-VII.1.
The SCALE documentation allows for application of covariances from a single library to many other cross-section libraries, because much of the uncertainty data is based on simplifying assumptions that do not depend on specific ENDF evaluations~\cite{scale}.
\begin{figure}[H]
    \centering
    \includegraphics[width=1.0\linewidth]{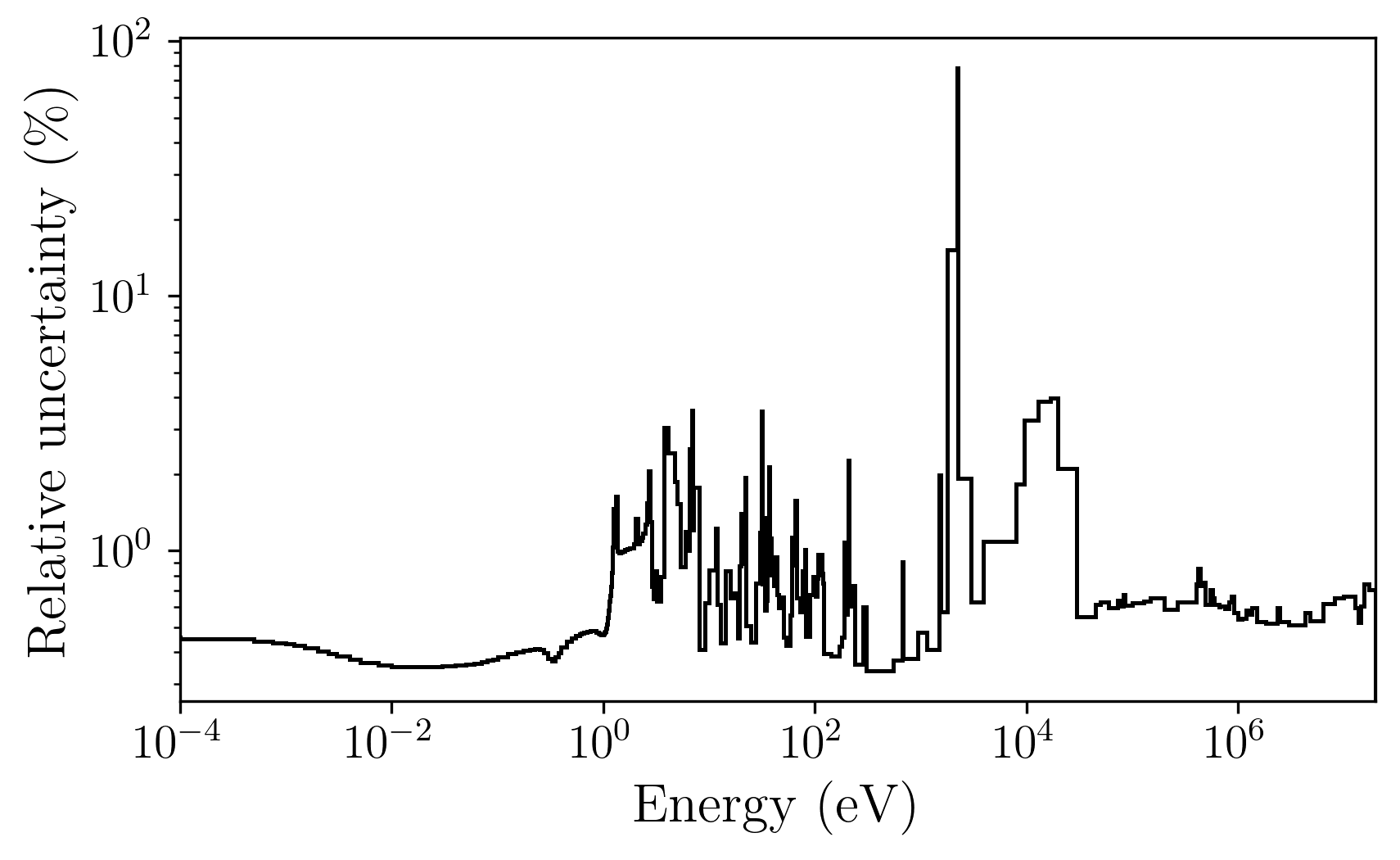}
    \caption{Relative uncertainty of the $^{235}\text{U}$ fission cross-section derived from the ENDF/B-VII.1 covariance library using the SCALE-252 energy group structure.}
    \label{fig:rel_uncertainties}
\end{figure}

 \subsection{Sensitivity Profile Calculation in OpenMC}
An eigenvalue sensitivity coefficient $S_{\sigma(r, E)}$ is a scalar quantity that describes the fractional change in $k_\text{eff}$ resulting from perturbations or uncertainties in nuclear data $\sigma(r, E)$. It is expressed as 
\begin{equation}
    S_{\sigma(r,E)} = \frac{\delta k_\text{eff}/k_\text{eff}}{\delta \sigma(r, E)/\sigma(r, E)}
\end{equation}
where $k_\text{eff}$ is the effective multiplication factor, $\sigma(r,E)$ is the cross section for reaction $r$ at energy $E$, and $\delta$ is the differential perturbation operation.
In this work, the nuclear data considered are exclusively cross sections.
The IFP method tracks neutrons across successive generations to weight their importance by the fission contribution of their progeny, and is used in this work to calculate sensitivity vectors within the SCALE-252 energy group structure.

Sensitivities were calculated using the Iterated Fission Probability (IFP) method~\cite{peng2017} on a non-production branch of OpenMC.
The OpenMC sensitivity calculation implementation was validated against the Serpent~\cite{serpent} Monte Carlo code to ensure accuracy, as discussed in Section~\ref{subsec:cask}.
The codes were found to be highly consistent in their calculated sensitivities.

\subsection{Surrogate Modeling}
\label{sec:surrmodel}
A surrogate model is a fast approximation of an expensive simulation that comes with several advantages at the cost of accuracy.
In this work, a deep neural network was trained as a surrogate model of OpenMC's sensitivity calculation capabilities.
Specifically, the neural network predicts the sensitivity profile $\hat S$ for a given experimental geometry $x$, where $x$ is encoded as an input tensor $x \in \mathbb{R}^{4\times32 \times 32}$ with each component representing the material assignment of that cell.
The predicted $\hat S$ is then used to compute $\hat c_k$ for the criticality experiment represented by $x$ with the target technology.
The model is trained to predict $\hat S$ rather than $\hat c_k$ directly in order to finely evaluate its performance across different reaction types and energy groups as shown in Section \ref{subsec:train}.
This approach also limits the complexity of what the model must learn to the most computationally expensive aspect of the problem, since the step of  calculating $\hat c_k$ from $\hat S$ is straightforward.
The \space$\hat{}$\space notation is used in this work to denote a quantity predicted by a surrogate model rather than calculated via Monte Carlo or other numerical or analytical method.

Using a surrogate neural network for sensitivity calculation has several advantages.
It is extremely fast to evaluate, requiring only a single forward pass on a GPU to calculate the full sensitivity profile for a geometry.
It is also differentiable, meaning that the gradient of a predicted quantity with respect to the input can be obtained.
In the context of experiment design, this means that the gradient of the predicted $\hat c_k$ with respect to input geometry $x$ can be obtained, and $x$ can then be directly perturbed using gradient descent in order to increase $\hat c_k$.
This approach is intractable using Monte Carlo sensitivity calculations.

Several methods are used to evaluate the accuracy of the model. 
For a single geometry, the Mean-Squared Error (MSE) is calculated as the average squared difference between each scalar sensitivity coefficient in the predicted sensitivity profile $\hat S$ and the OpenMC-calculated profile $S$, of which there are $R$ reactions and $G$ energy groups. 
The MSE is extended for a set of $N$ geometries such that (defining $\hat S^{(i)}_{r,g}$ and $S^{(i)}_{r,g}$ as the predicted and evaluated sensitivity coefficients of reaction $r$ at energy group $g$ for geometry $i$),
\begin{equation}
    \text{MSE} = \frac{1}{N}\sum_{i=1}^{N}\frac{1}{RG}\sum_{r=1}^{R}\sum_{g=1}^G(\hat{S}^{(i)}_{r,g} - S^{(i)}_{r,g})^2.
\end{equation}
In the context of design optimization, it is also useful to evaluate the Mean Absolute Error (MAE), calculated for $N$ geometries as
\begin{equation}
    \text{MAE} = \frac{1}{N}\sum_{i=1}^{N}\frac{1}{RG}\sum_{r=1}^{R}\sum_{g=1}^G|\hat{S}^{(i)}_{r,g} - S^{(i)}_{r,g}|.
\end{equation}

\subsection{Data Structure \& Generation}
\label{sec:ca}
For this work, a large number of experiment geometries were procedurally generated and used to build a dataset of OpenMC-calculated sensitivity profiles.
The OpenMC dataset generator models each cube with vacuum boundary conditions on all six external faces.
Reactivity and sensitivity are highly dependent on both the size of the configuration and on the overall mass proportion.
To control for this in examining the efficacy of the optimization method, the side length of all geometries is intentionally set at $200\,$cm, which is the maximum experiment size compatible with the proposed SPARC facility~\cite{woolstenhulme2025}.
The primary face of each cubic geometry is discretized into a 64x64 grid of materials, with quarter symmetry imposed. 
Each $3.125^2\,\text{cm}^2$ grid cell extends uniformly through the full depth of the cube to form a set of square prism voxels.
A palette of four materials was chosen representative of the target technology.
The palette included 20\% enriched $\text{UO}_2$, stainless steel, borated aluminum, and polyethylene.

Constraining critical experiment designs to a 2D grid of materials extruded through a cubic geometry has historical precedent in the field.
This approach was used by the ZPR~\cite{zpr} family of critical experiment facilities for 35 years to support highly flexible experiment designs.
Each facility contained an HST assembly comprised of a grid of nominally square $0.040$\,in thick stainless steel tubes, and grid dimensions ranged from 31x31 in the ZPR-3 machine to 77x77 in the expanded ZPPR-6 machine. 
The tubes would be filled with easily interchangeable plates of various materials to represent the composition of materials in the target reactor design.

\begin{figure}[H]
    \centering
    \includegraphics[width=0.75\linewidth]{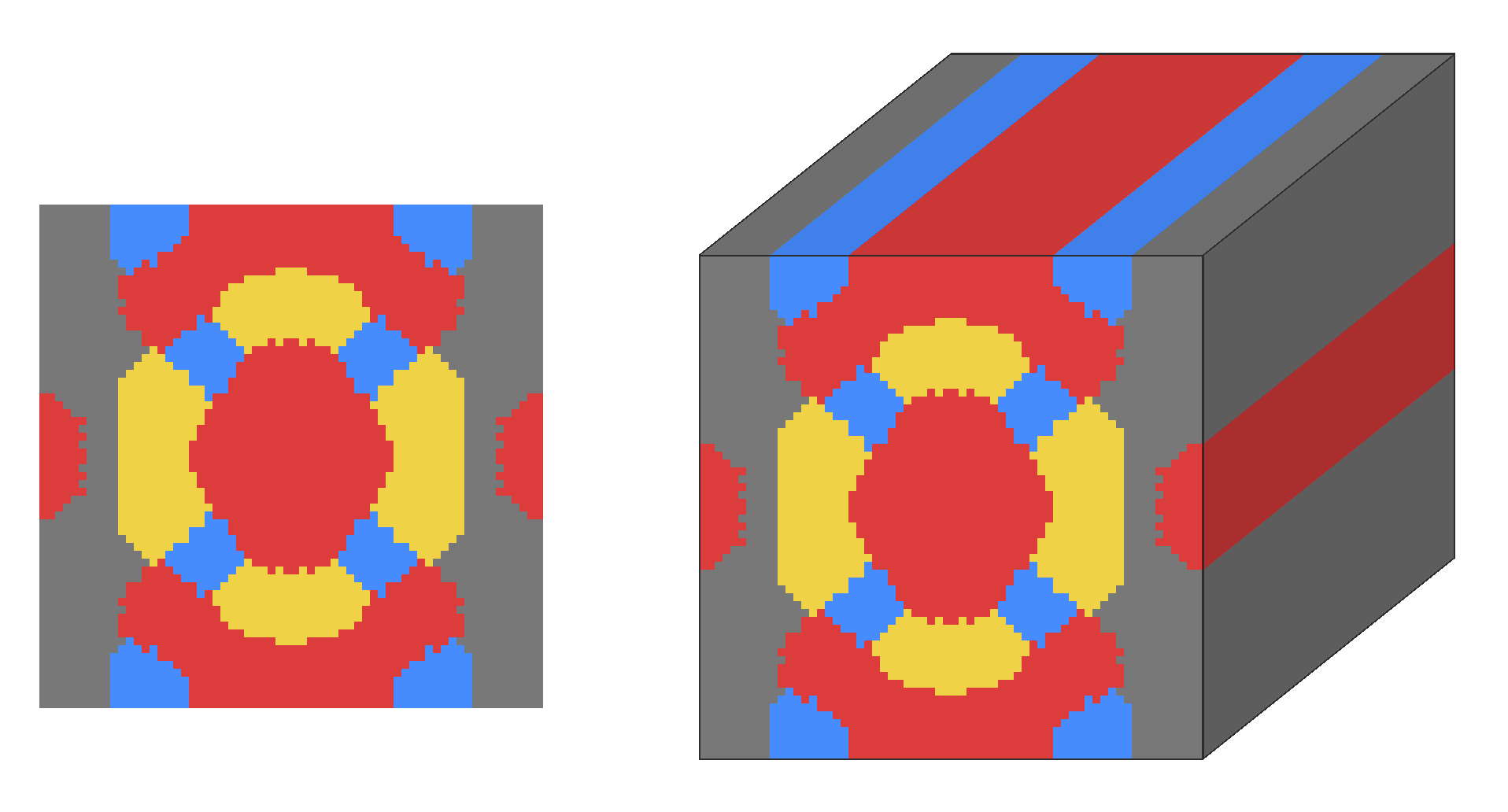}
    \caption{An example critical experiment geometry. The 64x64 material grid is extruded to form a cube with side lengths of $200\,\text{cm}$.}
    \label{fig:cube}
\end{figure}

To train the surrogate model, a large and diverse training dataset of example experiment configurations is required.
Different approaches to data generation are possible.
In this work, a cellular automaton algorithm  was used to procedurally generate experiment geometries.

\begin{figure}[H]
    \centering
    \includegraphics[width=1\linewidth]{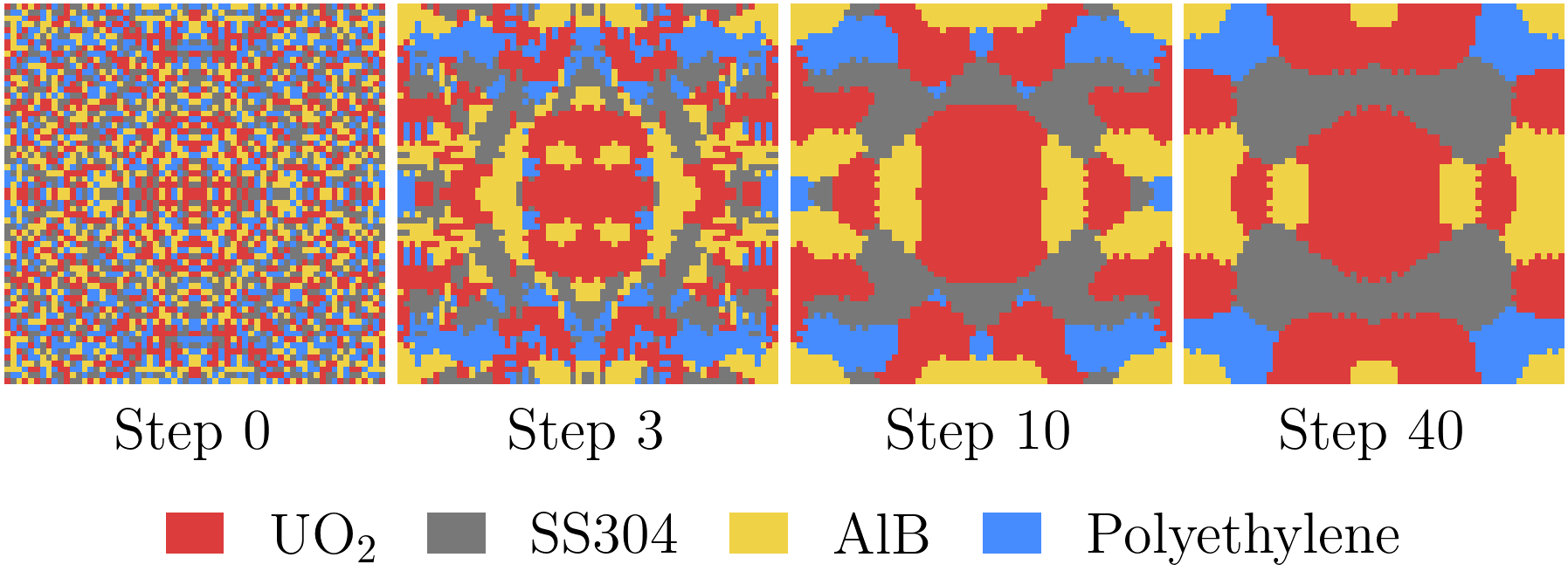}
    \caption{Procedural geometry generation over 40 cellular automaton steps.}
    \label{fig:ca_steps}
\end{figure}

\begin{figure}[H]
    \centering
    \includegraphics[width=1\linewidth]{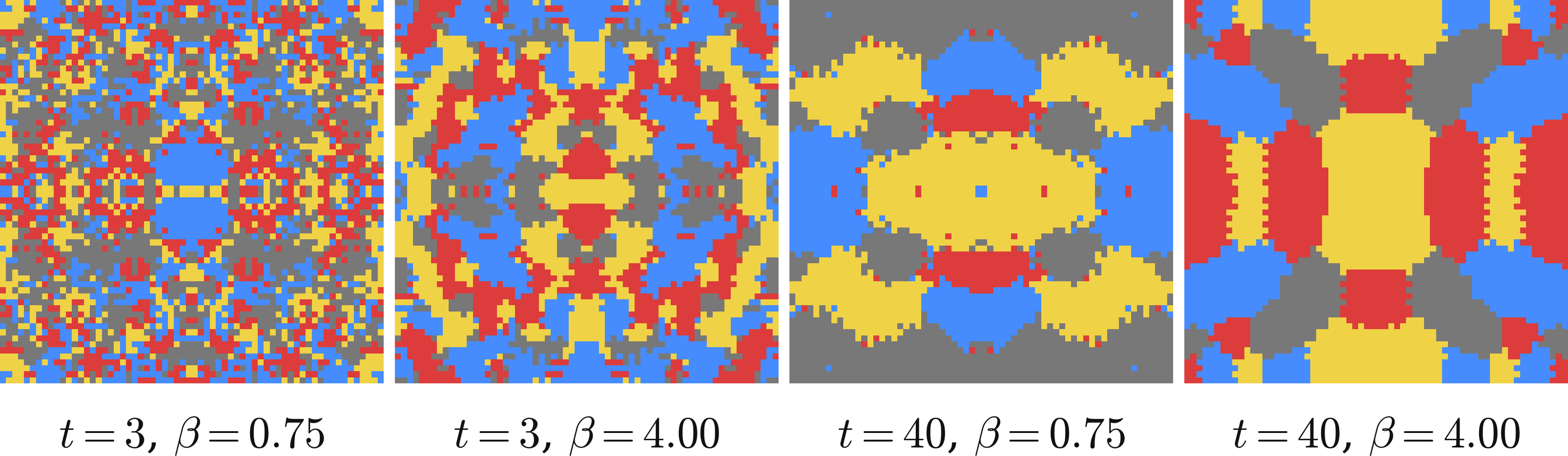}
    \caption{Geometries generated from the same initial seed with different combinations of $t$ and $\beta$.}
    \label{fig:ca_params}
\end{figure}

A cellular automaton is a process that describes a discrete grid of cells that update over time according to the state of their neighbors.
The geometry starts as a grid of cells randomly initiated by a unique seed.
The state of each cell is its material assignment, which is updated at each step by sampling from a distribution of its neighboring material counts, weighted by a factor $\beta$.
To encourage diverse results, the number of update steps $t$, and the weighting of neighboring materials $\beta$ are randomly chosen for each geometry that is generated.
Geometries generated with a higher $t$ feature larger, more homogeneous regions of material, since at each step it becomes more likely that a cell changes to the material of its neighbor.
Geometries generated with a higher $\beta$ feature sharper boundaries between regions, since the sampled probability distribution is much sharper. 
By varying these quantities throughout the dataset a wide distribution of geometries can be generated.
The full cellular automaton algorithm is given in Algorithm~\ref{alg:ca-generation}, and a comparison of geometries generated with different $t$ and $\beta$ is shown in Figure~\ref{fig:ca_params}.
It should be noted that the proposed experiment designs are not subject to this process, only the training data used to create the surrogate model.
The cellular automaton process is shown for an example geometry in Figure~\ref{fig:ca_steps}.

This approach was chosen for both the distribution of geometries it is able to generate, and for its natural scalability compared to other methods.
Because a cellular automaton is algorithmic, an arbitrary number of geometries can be generated without modifying the procedure.
Additionally, it emits a broad distribution of material combinations and structures within the geometries without requiring arbitrary human judgment to specify which structures should be included.
In this way, the dataset is able to represent a comprehensive design space of geometries without imposition of human design decisions, making it a suitable approach when exploring the preliminary performance of an unconstrained data driven optimization method.
A disadvantage of this approach is that without further constraints, it does not lend itself to the representation of geometries that are realistic or necessarily feasible.
The coverage of the resulting dataset is examined further in Section~\ref{subsec:train} using physically meaningful descriptors.

\begin{algorithm}[H]
\caption{Dataset Geometry Generation}
\label{alg:ca-generation}
\begin{algorithmic}[1]
\State $x \sim \text{Uniform}\{0, \ldots, 3\}^{32 \times 32}$ 
\For{$1,\ldots, t$}
    \For{each cell $(i, j)$}
        \State Update neighbor\_counts
        \State $x'[i,j] \sim \text{softmax}(\beta\cdot\text{neighbor\_counts})$
    \EndFor
    \State $x \leftarrow x'$
\EndFor
\State Reflect $x$ to the full $64 \times 64$ grid
\State \Return $x$
\end{algorithmic}
\end{algorithm}

\subsection{Network Architecture}

The deep neural network used as a surrogate model is not comprised of simple multilayer perceptrons.
Instead, it uses a heterogeneous architecture to more effectively encode the relationships between the experimental geometry and the sensitivities.
Model architectural design was guided by both the spatial nature of the input and certain aspects of the underlying transport physics.

A geometry encoding $x$ is passed through a U-Net~\cite{unet} to produce a spatial feature map, which is then processed by a multigroup attention pooling layer (introduced in this work) that weights features by their spatial importance for each discretized energy group.
The resulting group descriptors are passed through a 1D convolutional regressor operating over the full energy group structure to produce a predicted sensitivity vector $\hat s$ for each reaction.
The sensitivity vectors are combined to create the final predicted sensitivity profile $\hat S$.
\begin{figure*}[tb]
    \centering
    \includegraphics[width=\textwidth]{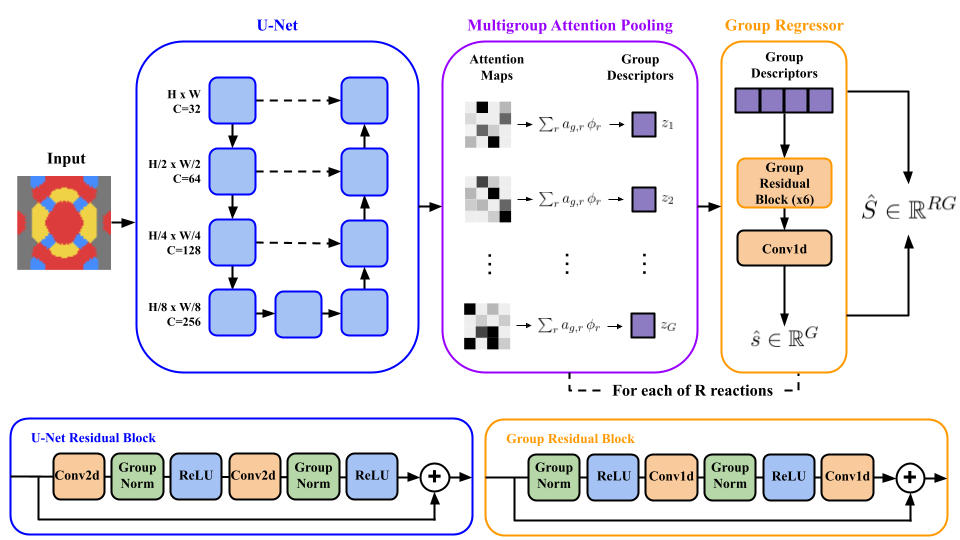}
    \caption{Surrogate model architecture. The experiment geometry is passed through a shared U-Net to extract spatial features. The features are weighted by isolated attention pools for each energy group and reaction, regressed into the scalar sensitivity coefficients, and combined to create the final predicted sensitivity profile $\hat S \in \mathbb{R}^{RG}$ for $R$ unique reactions and $G$ discrete energy groups.}
    \label{fig:arch}
\end{figure*}
\subsubsection{U-Net Encoder}
The U-Net is an encoder-decoder, fully convolutional network architecture first introduced in 2015 for biomedical image segmentation~\cite{unet}. 
The defining feature of the U-Net is its use of skip-connections, which transfer information from earlier, higher-resolution feature maps directly to the decoder, allowing the network to recover fine spatial detail that would otherwise be lost during downsampling.
The encoder consists of four successive modules comprising a convolution, group normalization, ReLU activation, and residual block, followed by a max pooling layer between each stage.
The residual blocks consist of two group-normalized convolutions with ReLU activations.
The decoder mirrors this structure with transposed convolutions for upsampling, concatenated with the corresponding encoder skip connection.
The output of the decoder is a feature map $\Phi \in \mathbb{R}^{C \times H \times W}$ where $C$ is the number of channels in the first encoder stage, and $H \times W$ is the original input resolution.

\subsubsection{Multigroup Attention Pooling}
\label{sec:map}
Let $\phi_r \in \mathbb{R}^C$ denote the feature vector extracted by the U-Net encoder at spatial position $r \in \{1, \dots, HW\}$. For each energy group $g \in \{1,\dots,G\}$, a learned attention vector $w_g \in \mathbb{R}^C$ produces a scalar importance score $w_g^\top \phi_r$ at every spatial location. These scores are normalized via softmax to yield $a_g \in \mathbb{R}^{HW}$, a spatial attention map over the geometry,
\begin{equation}
    a_{g,r} = \frac{\exp(w_g^\top \phi_r)}{\sum_{r'=1}^{HW} \exp(w_g^\top \phi_{r'})}.
\end{equation}
Each $a_g$ records the regions of the geometry that are most informative for energy group $g$. 
Because neutrons with different energies interact with geometry at different characteristic length scales, independently learned attention maps per group provide a physically motivated inductive bias for the model to learn.
The group descriptor $z_g$ is then taken as the attention-weighted sum of the spatial feature vectors,
\begin{equation}
    z_g = \sum_{r=1}^{HW} a_{g,r}\, \phi_r \;\in\; \mathbb{R}^C.
\end{equation}
Stacking over all $G$ groups gives the descriptor matrix $Z \in \mathbb{R}^{G \times C}$.

\subsubsection{Group Regressor}
The group descriptors $Z \in \mathbb{R}^{G \times C}$ are treated as a 1D sequence of length $G$ and processed by a convolutional residual network that maps each group to a final scalar sensitivity coefficient.
The residual network is formed by six successive blocks comprised of a group normalization, ReLU activation, and 1D convolution, with a residual connection after every other block.
The output of the residual network is projected through a final 1D convolution into the predicted sensitivity vector $\hat s \in \mathbb{R}^G$ for each reaction.

When trained on $R > 1$ reactions, the U-Net encoder weights are shared across all reactions, but each reaction retains independent attention pool and regressor weights.
The $R$ sensitivity vectors are flattened to form the full sensitivity profile $\hat S \in \mathbb{R}^{RG}$.

\subsubsection{Training}
A training dataset was generated using the cellular automata algorithm described in Section~\ref{sec:ca} with parameters $t \in [3, 40]$ and $\beta \in [0.75, 4]$.
Each profile contained $k_\text{eff}$ sensitivities to the cross-sections of $^{235}\text{U}$ fission, $^1\text{H}$ elastic scattering, $^{12}\text{C}$ elastic scattering, and $^{58}\text{Ni}$ absorption.
Sensitivities were calculated via the IFP method in OpenMC using 25000 particles, 20 inactive generations, 250 active generations, and 3 latent IFP generations.
These simulation parameters were chosen to ensure an average dataset uncertainty of less than $50\,$pcm.
However, a disadvantage of this method is that training data uncertainties are not persisted or accounted for in later prediction of $c_k$.
This improvement should be a target of future work.

Sensitivities to different reactions and energies span a wide range of magnitudes. 
To encourage all sensitivity coefficients to be treated with equal importance during training, the model was trained on z-scores of normalized sensitivities using a standard mean squared error (MSE) loss function, with the mean and standard deviation of each reaction saved in the model and recovered during inference.

The AdamW~\cite{adamw} optimizer is used to train the model end-to-end by minimizing the mean squared error between the predicted and target sensitivity coefficients, normalized to $\mathcal{N}(0, 1)$, such that the training loss is
\begin{equation}
    \mathcal{L} = \frac{1}{N}\sum_{i=1}^{N}\frac{1}{RG}\sum_{r=1}^{R}\sum_{g=1}^G(\hat{S}^{(i)}_{r,g} - S^{(i)}_{r,g})^2
\end{equation}
where $\hat S_{r,g}$ and $S_{r,g}$ are the predicted and normalized target sensitivity coefficients respectively for reaction $r$ and energy group $g$.

Since the focus of this work is the method and application of a novel model and task, rather than improvement upon an existing benchmark or neural network architecture, no structured hyperparameter sweep was run during training.
To validate the inclusion of the attention pooling layer, versions of the model were trained with different choices of pooling layer and compared in Table~\ref{tab:ablation} and Figure~\ref{fig:loss_curves}.
As discussed further in Section~\ref{subsec:train}, multigroup attention pooling demonstrated the lowest error of the benchmarked options under the chosen training hyperparameters, justifying its inclusion in the model.
This result indicates the potential value of custom neural network architectures for modeling neutronics, which is explored further in Section~\ref{subsec:interp}.

The full architecture details and training hyperparameters for the surrogate model are given in Table~\ref{tab:hparams} in~\ref{sec:app_hparams}.
\subsection{Gradient Based Optimization}
\label{sec:gbo}
The design of an experiment geometry to maximize $c_k$ with a target technology can be mathematically framed as a single-objective optimization problem.
In general, a single-objective optimization problem can be expressed as
\begin{equation}
\begin{aligned}
\min_{x \in \mathbb{R}^n} \quad & \mathcal{L}(x) \\
\text{subject to} \quad 
& g_i(x) \le 0, \quad i = 1,2,\dots,n_g, \\
& h_j(x) = 0, \quad j = 1,2,\dots,n_h .
\end{aligned}
\end{equation}

Let $f_\theta$ be the surrogate neural network, such that $\hat S=f_\theta(x)$ is the predicted sensitivity profile for input geometry $x$. For this application, the objective function $\mathcal{L}(x)=-\hat c_k$ such that the optimization goal is to produce an experiment geometry $x$ that maximizes $c_k$ with the target technology of interest.
The $\hat c_k$ notation is introduced to differentiate between the neural-network predicted $\hat c_k$ and the $c_k$ calculated via Monte Carlo.
The application is formulated such that there are no applicable constraints on $x$, as any arrangement of materials in the grid produces a valid sensitivity profile.
It can therefore be thought of that $n_g=n_h=0$.
However, this optimization is within the parameters of the design space of $x$. That is, all possible inputs are constrained to 64x64 material grids within cubic geometries with $200\,$cm side lengths.
This then presents a distinction between the free and fixed parameters of optimization at a higher level: the parameters of size, material options, and grid fidelity are fixed, and the parameters of material choices within the grid are free.

An important shortcoming of this study in its current form is the lack of consideration for the actual $k_{\text{eff}}$ of the criticality experiment.
The constraint that $k_{\text{eff}} > 1$ for compatibility with the HST experiment approach should realistically be accounted for, as well as more nuanced constraints to ensure the physical feasibility of experiments.

The surrogate \(f_\theta\) is differentiable with respect to $x$, meaning that the gradient of \(\hat c_k\) with respect to the input geometry can be computed by backpropagation through the model.
The candidate geometry \(x\) is encoded as one-hot material assignments, meaning that over each spatial position in the grid, the component for the assigned material is 1, and the rest 0.
This is not immediately differentiable.
To allow for gradient-based optimization, the problem is relaxed into optimization over continuous material logits \(\ell \in \mathbb{R}^{4 \times 32 \times 32}\), where the four material channels correspond to the four possible material assignments and the \(32\times32\) spatial dimensions correspond to the quarter-grid design domain.
Before evaluating the surrogate, the quarter-grid logits are reflected across both spatial axes to form full-grid logits on a \(64\times64\) grid.

Given $M$ possible materials, the probability \(p\) of assigning material \(m\) to position \(r\) is obtained by taking a temperature-scaled softmax over the material logits
\begin{equation}
    p_{m,r}
    =
    \frac{\exp(\ell_{m,r}/\tau)}
    {\sum_{m'=1}^{M}\exp(\ell_{m',r}/\tau)}
\end{equation}
where \(\tau\) is the softmax temperature.
For optimization steps \(t=0,\ldots,T-1\), the temperature is held fixed at \(\tau=1\) for the first third of the optimization to encourage exploration, and then cosine-annealed to a minimum value of \(0.2\).
The temperature schedule is then
\begin{equation}
    \tau_t =
    \begin{cases}
    1, & t<\lfloor T/3\rfloor\\[4pt]
    0.2 + 0.4(1+\cos(\pi\cdot\alpha_t)) & t\ge \lfloor T/3\rfloor
    \end{cases}
\end{equation}
with
\begin{equation}
    \alpha_t = \frac{t-\lfloor T/3\rfloor}{\max(1,T-\lfloor T/3\rfloor-1)}
\end{equation}
such that as \(\tau\) decreases, the relaxed probabilities become increasingly concentrated on the material with the largest logit, so that in the limit
\begin{equation}
    \lim_{\tau \to 0} p_{m,r}
    =
    {1}\!\left[
    m=\arg\max_{
    m'} \ell_{m',r}
    \right]
\end{equation}
where $1[\cdot]$ denotes an indicator function, such that the expression approaches 1 for the selected material and 0 for all other materials. This allows the softmax to remain differentiable for any finite \(\tau>0\), while becoming increasingly faithful to the original discrete material-assignment problem as \(\tau\) decreases.

At each optimization step, the surrogate must be evaluated on a discrete geometry, so the forward-pass geometry \(x\) is formed by taking the most probable material at each spatial position
\begin{equation}
    x_{m,r}
    =
    1\!\left[
    m=\arg\max_{m'} p_{m',r}
    \right].
\end{equation}
The tensor passed to the surrogate is implemented using a straight-through estimator (STE)~\cite{bengio}
\begin{equation}
    x_{\mathrm{STE}} = x + p - \mathrm{stopgrad}(p).
\end{equation}
In the forward pass, \(x_{\mathrm{STE}}=x\), so the surrogate is evaluated on a hard one-hot geometry.
In the backward pass, the hard assignment is treated as the identity map with respect to \(p\), allowing gradients to flow through the relaxed probabilities and then through the softmax to the logits, so that
\begin{equation}
\label{eq:ste}
    \frac{\partial x_{\mathrm{STE}}}{\partial p} \approx I.
\end{equation}
The surrogate prediction is denormalized before computing the correlation coefficient.
If \(f_\theta(x_{\mathrm{STE}})\) denotes the normalized surrogate output, then the predicted sensitivity is
\begin{equation}
    \hat S
    =
    \sigma \odot f_\theta(x_{\mathrm{STE}}) + \mu
\end{equation}
where \(\mu\) and \(\sigma\) are the normalization statistics stored with the surrogate checkpoint.
The predicted correlation coefficient  $\hat c_k$ is then computed with Equation \ref{eq:ckdefinition}.

The optimization objective is to maximize $\hat c_k$, such that the minimized loss is
\begin{equation}
    \mathcal{L}
    =
    -\hat c_k
\end{equation}
Using the straight-through estimator, the gradient of the loss with respect to the material logits is
\begin{equation}
    \nabla_\ell \mathcal{L}
    =
    \frac{\partial \mathcal{L}}{\partial \ell}
    =
    -\frac{\partial \hat c_k}{\partial \hat S}
    \frac{\partial \hat S}{\partial f_\theta}
    \frac{\partial f_\theta}{\partial x_{\mathrm{STE}}}
    \frac{\partial x_{\mathrm{STE}}}{\partial p}
    \frac{\partial p}{\partial \ell}
\end{equation}
Since the STE replaces \(\partial x_{\mathrm{STE}}/\partial p\) by the identity as shown in Equation~\ref{eq:ste}, this becomes
\begin{equation}
    \nabla_\ell \mathcal{L}
    =
    \frac{\partial \mathcal{L}}{\partial \ell}
    \approx
    -\frac{\partial \hat c_k}{\partial \hat S}
    \frac{\partial \hat S}{\partial f_\theta}
    \frac{\partial f_\theta}{\partial x_{\mathrm{STE}}}
    \frac{\partial p}{\partial \ell}
\end{equation}
This formulation is fully differentiable with respect to the logits and can be treated with gradient descent.
In practice, the logits are updated with the AdamW~\cite{adamw} optimizer, such that to update from step $t$ to step $t+1$ is
\begin{equation}
    \ell_{t+1}
    =
    \mathrm{AdamW}
    \left(
    \ell_t,\nabla_{\ell_t}\mathcal{L}_t
    \right)
\end{equation}
Minimizing the loss in this way is equivalent to maximizing $\hat c_k$ through gradient ascent.

Upon convergence, a discrete geometry is recovered by assigning each position \(r\) the material with the highest logit value \(\ell_{m,r}\).
The resulting \(32\times32\) quarter-grid material assignment is reflected to form a \(64\times64\) full-grid geometry.
This full-grid geometry is evaluated in OpenMC to produce a final $c_k$.
The full procedure is given in Algorithm~\ref{alg:gbo} for a single initialization.

In practice, this procedure is applied to many independently initialized candidate geometries in parallel to maximize coverage of the loss landscape from many different starting positions.
A batch of $N$ geometries are initialized and optimized in parallel.
Upon convergence, the geometries with the top $k$ predicted correlations are evaluated in OpenMC, and the geometry with the highest true $c_k$ is returned.
\begin{algorithm}[H]
\caption{Gradient-Based Geometry Optimization}
\label{alg:gbo}
\begin{algorithmic}[1]
\State Initialize logits $\ell \in \mathbb{R}^{4 \times 32 \times 32},\,\,\,\ell_r \sim \mathcal{N}(0,1)\ \forall r$ 
\For{\(t = 0, \ldots, T-1\)}
    \State Compute \(\tau_t\) using the annealing schedule
    \State Reflect \(\ell_t\) to the full \(64\times64\) grid
    \State \(p_t \leftarrow \mathrm{softmax}(\ell_t/\tau_t)\)
    \State \(x_t \leftarrow \mathrm{onehot}(\arg\max_m p_{t,m})\)
    \State \(x_{\mathrm{STE},t} \leftarrow x_t + p_t - \mathrm{stopgrad}(p_t)\)
    \State \(\hat S_t \leftarrow \sigma \odot f_\theta(x_{\mathrm{STE},t}) + \mu\)
    \State \(\hat c_{k,t} \leftarrow
    \dfrac{
    S_{\mathrm{TARGET}}^\top C_{XX}\hat S_t
    }{
    \sqrt{
    \left(S_{\mathrm{TARGET}}^\top C_{XX}S_{\mathrm{TARGET}}\right)
    \left(\hat S_t^\top C_{XX}\hat S_t\right)
    }
    }\)
    \State \(\mathcal{L}_t \leftarrow -\hat c_{k,t}\)
    \State \(\ell_{t+1} \leftarrow \mathrm{AdamW}(\ell_t, \nabla_{\ell_t}\mathcal{L}_t)\)
\EndFor
\State \(x \leftarrow \mathrm{onehot}(\arg\max_m \ell_m)\)
\State Reflect $x$ to the full \(64\times64\) grid
\State \Return $x$ and OpenMC-computed $c_k$
\end{algorithmic}
\end{algorithm}

\section{Results and Discussion}
\label{sec:results}
\subsection{TN-LC Configurations and Sensitivities}
\label{subsec:cask}
In order to verify the models used for sensitivity calculation, OpenMC and Serpent were used to calculate $k_\text{eff}$ for the Dry Cask, Case 1 (flooded cask, low areal poison density), and Case 2 (flooded cask, high areal poison density) TN-LC configurations.
Parameters and resulting $k_\text{eff}$ values are shown in Table~\ref{tab:models}.
\begin{table*}[tb]
\centering
\renewcommand{\arraystretch}{1.15}
\begin{tabular*}{\textwidth}{@{\extracolsep{\fill}}lccc}
\hline
 & \textbf{Dry Cask} & \textbf{Case 1} & \textbf{Case 2} \\
\hline
Foam Density               & N/A                         & $0.12\text{ g/cm}^3$        & $0.22\text{ g/cm}^3$        \\
$^{10}$B Areal Density     & $0.0007\text{ g/cm}^2$      & $0.0007\text{ g/cm}^2$      & $0.0702\text{ g/cm}^2$      \\
H$_2$O VF                  & $0.0$                       & $0.8$                       & $0.8$                       \\
OpenMC $k_\text{eff}$      & $1.08976 \pm 16\text{ pcm}$ & $1.19240 \pm 21\text{ pcm}$ & $0.92146 \pm 22\text{ pcm}$ \\
Serpent $k_\text{eff}$     & $1.08832 \pm 3\text{ pcm}$  & $1.19275 \pm 5\text{ pcm}$  & $0.92061 \pm 7\text {pcm}$  \\                   
\hline
\end{tabular*}
\caption{Comparison of TN-LC model configurations.}
\label{tab:models}
\end{table*}
To benchmark the OpenMC sensitivity calculation implementation, the three TN-LC configurations were also modeled in the Serpent~\cite{serpent} Monte Carlo code.

\begin{figure*}[t]
    \centering
    \includegraphics[width=1\linewidth]{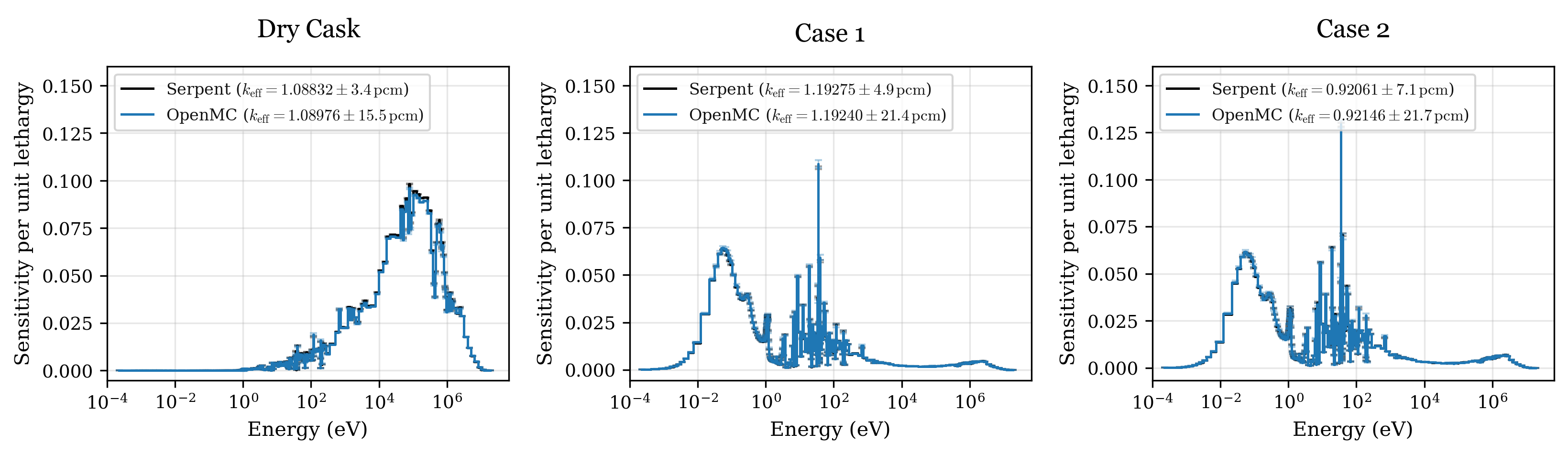}
    \caption{$^{235}$U sensitivity vectors calculated in both Serpent and OpenMC for the three TN-LC configurations.}
    \label{fig:serpent}
\end{figure*}

Sensitivities in both codes were calculated using 500,000 particles over 500 active generations and 50 inactive generations.
Figure~\ref{fig:serpent} shows the close agreement $^{235}$U fission sensitivity profiles calculated by OpenMC and Serpent.
While some bias is visible between the two codes, both used the ENDF/B-VIII.0 nuclear data library.
It is possible that this bias is due to a difference in the handling of boundary conditions between the codes.
Nonetheless, the consistency is enough to provide confidence in the sensitivity profiles used throughout the remainder of this analysis.

\subsection{Surrogate Dataset and Training}
\label{subsec:train}
\begin{figure}[H]
    \centering
    \includegraphics[width=0.9\linewidth]{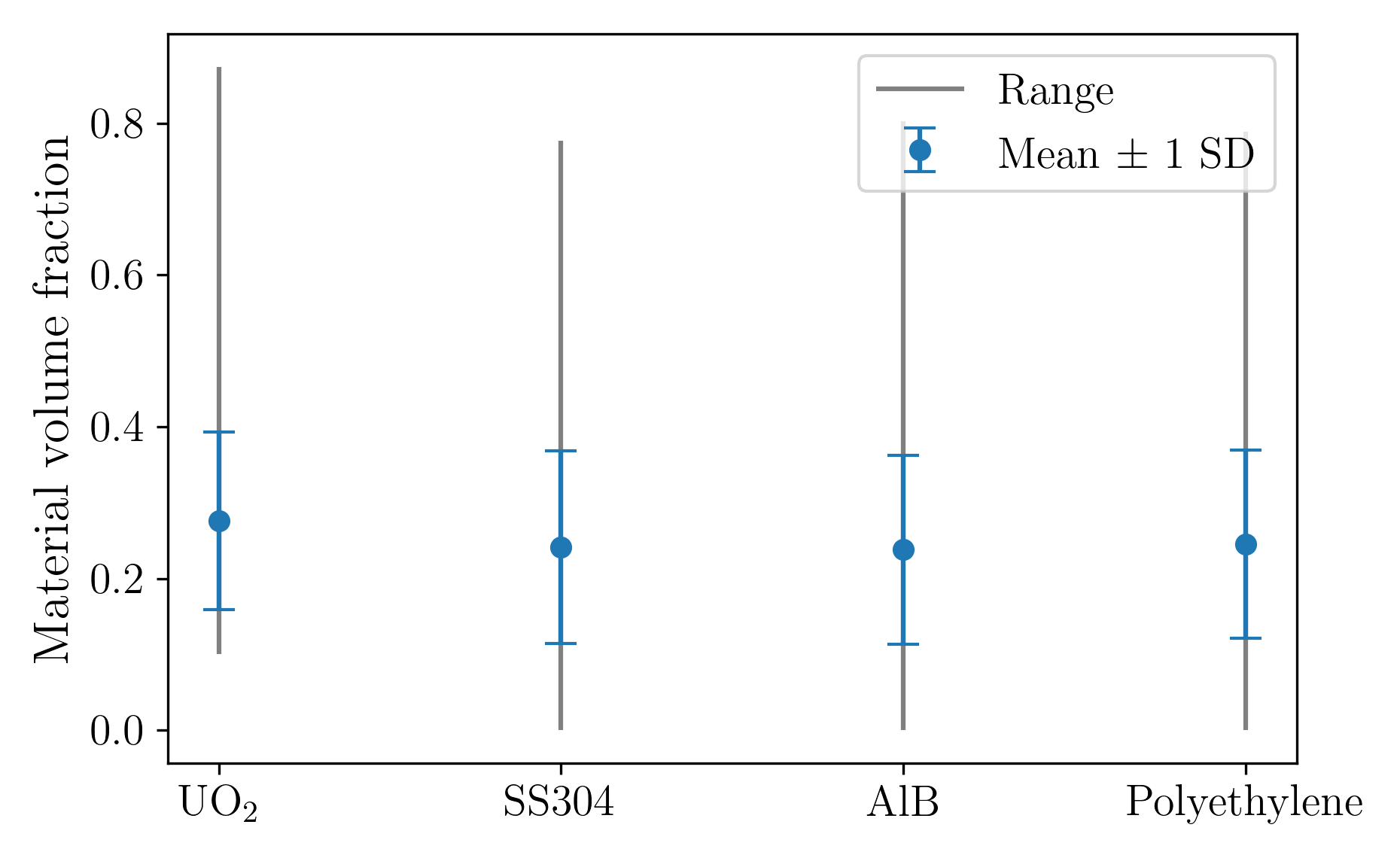}
    \caption{Mean ($\pm 1 \sigma$) and range of volume fractions for each material in the 3,000 training geometries.}
    \label{fig:diversity}
\end{figure}

\begin{figure}[H]
    \centering
    \includegraphics[width=0.9\linewidth]{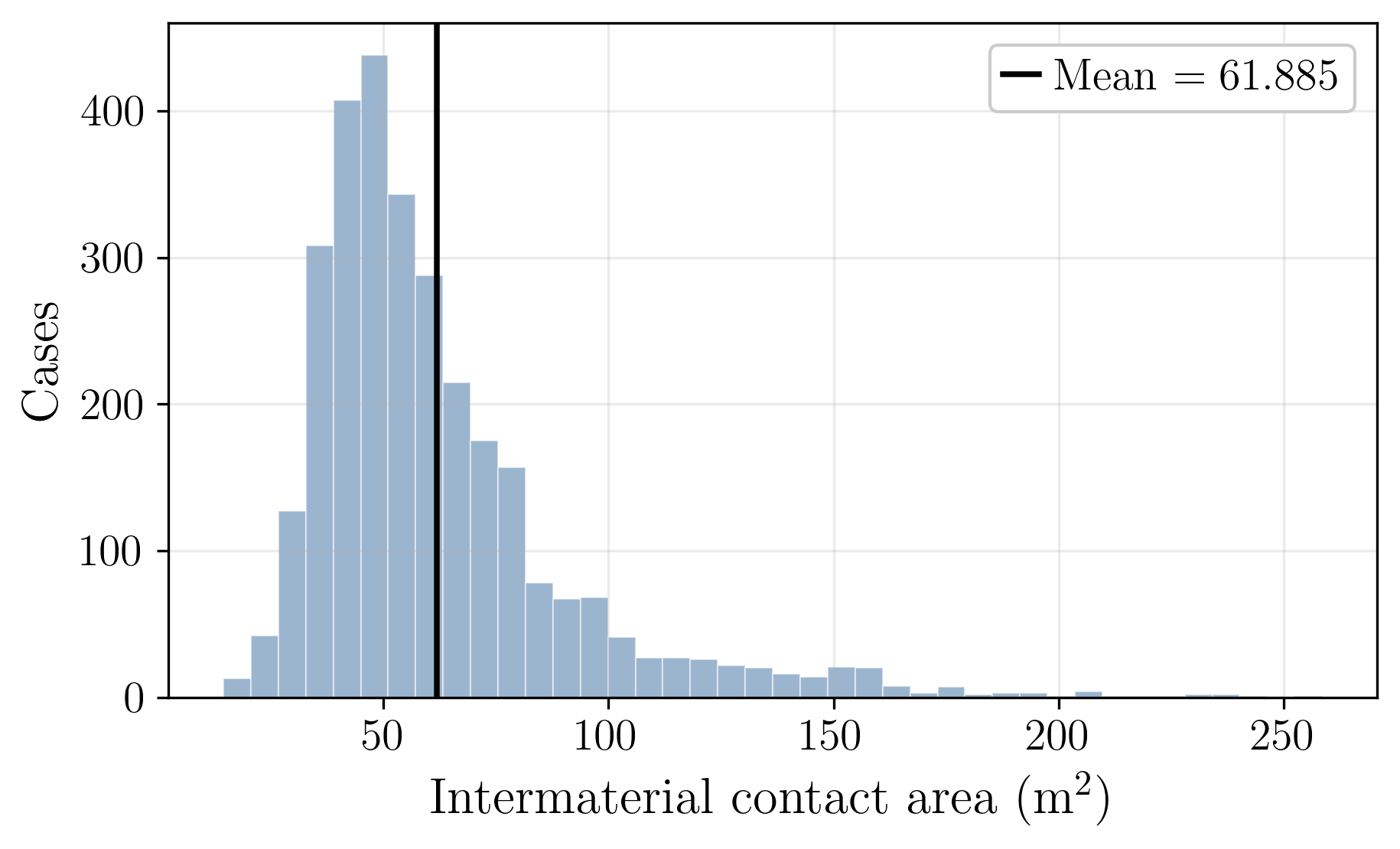}
    \caption{Distribution of the intermaterial contact area across the 3,000 training geometries, a measure of material heterogeneity.}
    \label{fig:contact}
\end{figure}

\begin{figure}[H]
    \centering
    \includegraphics[width=0.9\linewidth]{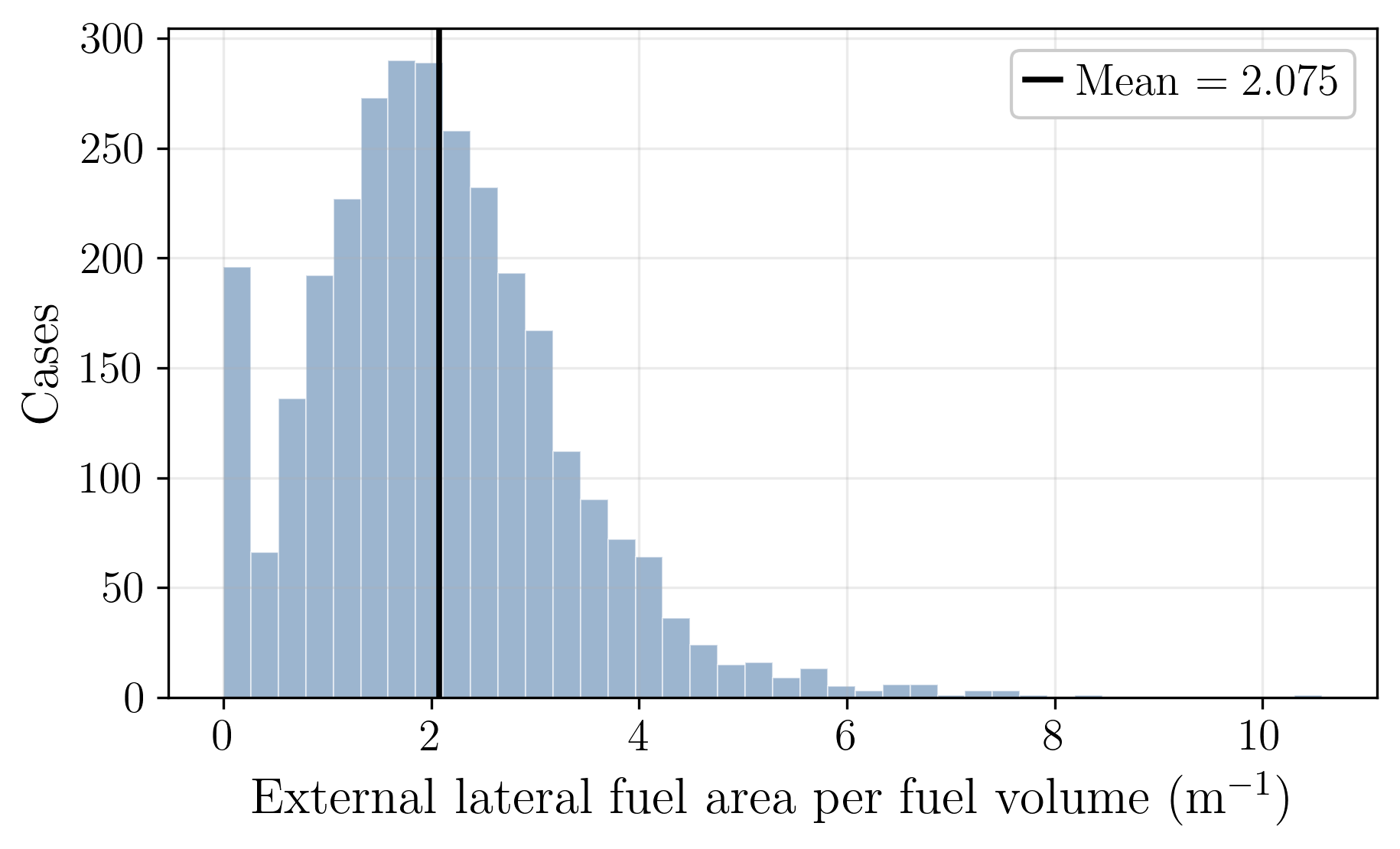}
    \caption{Distribution of a geometric proxy for leakage across the 3,000 training geometries.}
    \label{fig:leakage}
\end{figure}

\begin{figure*}[tb]
    \centering
    \includegraphics[width=\textwidth]{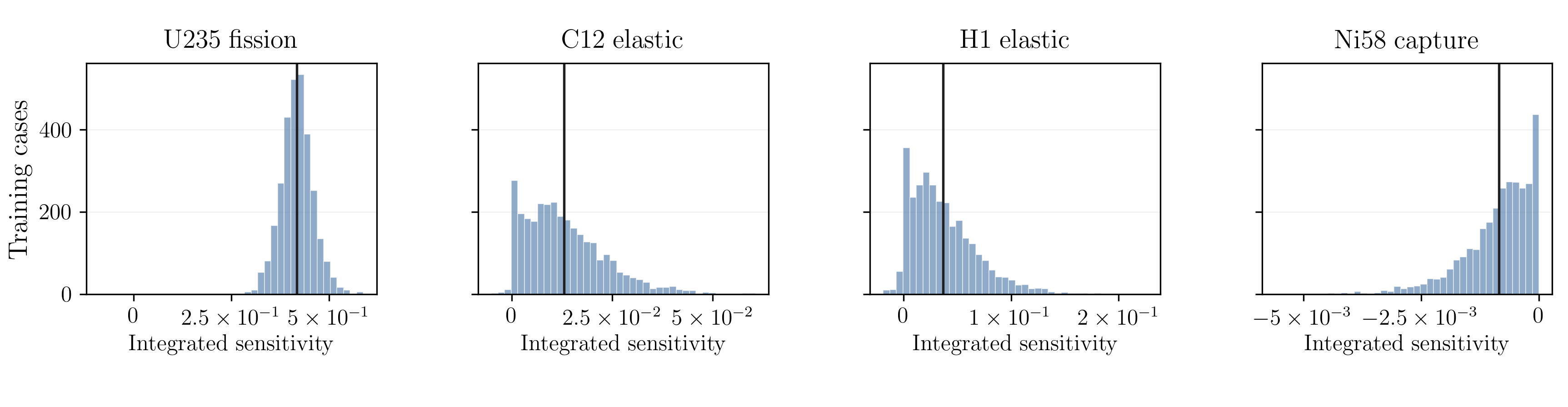}
    \caption{Distribution and mean of the integrated $k_\text{eff}$ sensitivity to each reaction across the 3,000 training geometries.}
    \label{fig:int_sens_data}
\end{figure*}

\begin{figure}[H]
    \centering
    \includegraphics[width=0.9\linewidth]{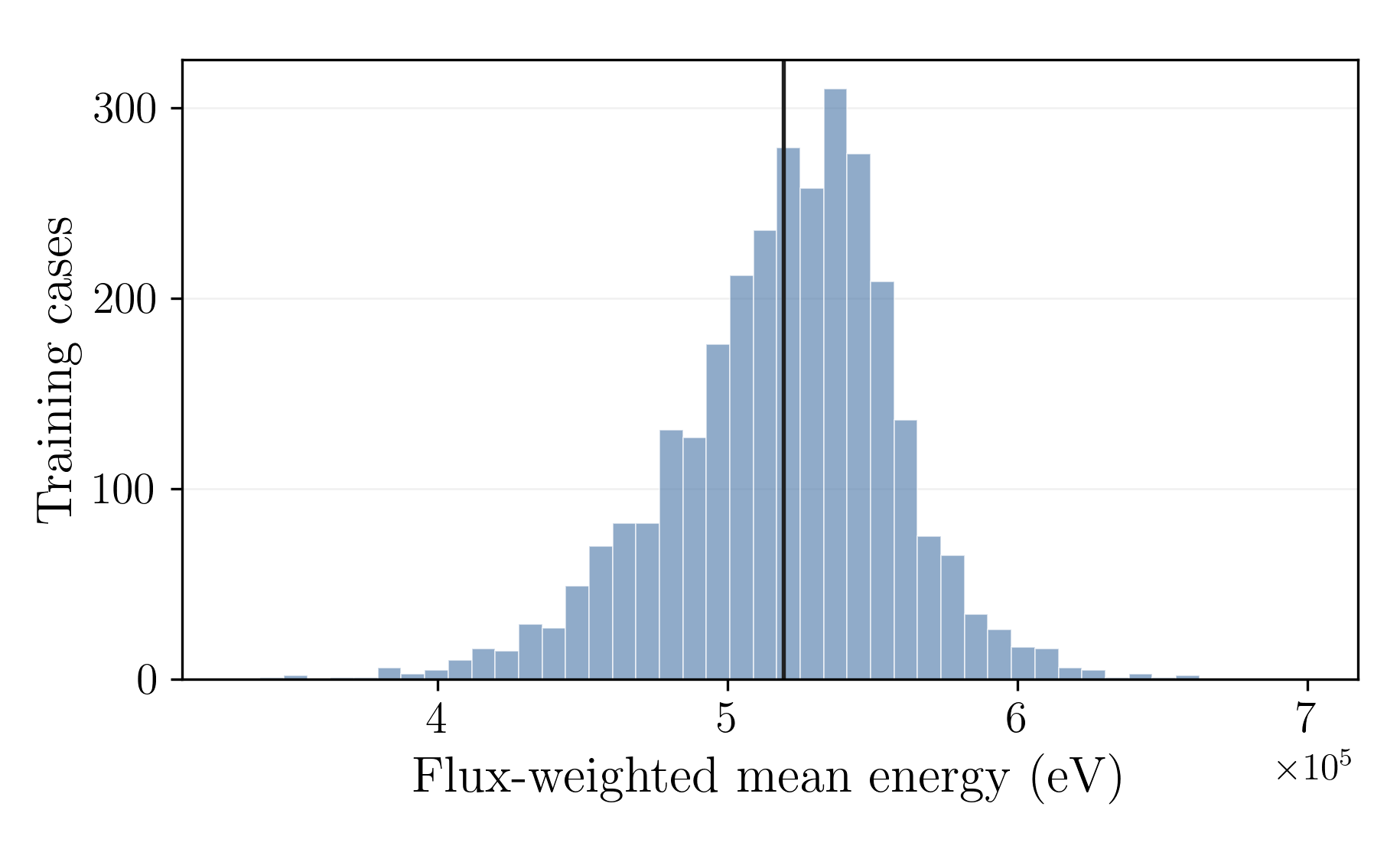}
    \caption{Distribution and mean of the flux-weighted mean neutron energy $\bar E$ across the 3,000 training geometries.}
    \label{fig:flux_energy_data}
\end{figure}

Sensitivities were calculated in OpenMC for a dataset of 5,000 experiment geometries generated using the cellular automata algorithm described in Algorithm~\ref{alg:ca-generation}, with average $k_\text{eff}$ uncertainty of about $40\,\text{pcm}$.
As a measure of dataset diversity, the distributions of material volume fractions across the 3,000 training geometries are shown in figure~\ref{fig:diversity}.
The geometries in the dataset are shown to span a wide distribution of material ratios.
\begingroup{}
The distribution of the intermaterial contact area, defined as the total area of the interfaces between different materials, is also shown in Figure~\ref{fig:contact} as a measure of material heterogeneity and fuel distribution in the geometries.
The geometric leakage proxy, defined as the area of fuel on the external lateral faces of the geometry per unit fuel volume, is shown in Figure~\ref{fig:leakage} as a measure of the proximity of the fuel to the vacuum boundaries.
However, two configurations with similar material ratios may exhibit very different neutronic behavior, so these geometric measures alone do not characterize the physical coverage of the design space.
To better characterize the dataset, two physically meaningful neutronic descriptors were calculated in OpenMC for each of the 3,000 geometries in the training set.

The first descriptor is the integrated sensitivity of $k_\text{eff}$ to each reaction, which is calculated as
\begin{equation}
    I_r = \sum_{g=1}^{G} S_{r,g}\,\Delta u_g
\end{equation}
where $S_{r,g}$ is the sensitivity per unit lethargy of reaction $r$ in energy group $g$, and $\Delta u_g$ is the lethargy width of the group.
The integrated $^{235}$U fission sensitivity describes the importance of fission in the fuel.
The integrated $^{1}$H and $^{12}$C elastic scattering sensitivities describe the moderation conditions provided by the polyethylene.
The integrated $^{58}$Ni capture sensitivity describes absorption in the stainless steel.
The distribution of each integrated sensitivity across the training set is shown in Figure~\ref{fig:int_sens_data}.

The second descriptor is the flux-weighted mean neutron energy $\bar E$, which is used as a spectral indicator and is calculated as
\begin{equation}
    \bar E = \frac{\int E\,\phi(E)\,dE}{\int \phi(E)\,dE}
\end{equation}
where $\phi(E)$ is the scalar flux over the entire geometry.
Both integrals are tallied directly in OpenMC over the energy range of the SCALE-252 group structure.
The distribution of $\bar E$ across the training set is shown in Figure~\ref{fig:flux_energy_data}.

As discussed in Section~\ref{subsec:app}, the TN-LC application cases lie outside the training data for several of these descriptors.
\endgroup{}

The performance of the surrogate model is evaluated by comparing predicted sensitivity profiles to OpenMC-evaluated ground truths.
Because the surrogate is trained only on grid-based experiment geometries, it cannot be evaluated on geometries that do not fit that constraint, such as comparison to experimental results from a source such as the International Criticality Safety Benchmark Evaluation Project handbook (ICSBEP)~\cite{icsbep}.
As defined in Section \ref{sec:surrmodel}, mean squared error (MSE) and mean absolute error (MAE) are used as metrics for surrogate accuracy.
Test and validation sets of 1,000 geometries were held fixed, and the model was iteratively trained with increasingly large training sets until the test sensitivity error fell beneath $50\,$pcm and began to plateau. 
This point was found to occur once the size of the training set reached 3,000 geometries, at which point the training dataset was held constant for further evaluations.
\begin{figure}[H]
    \centering
    \includegraphics[width=0.9\linewidth]{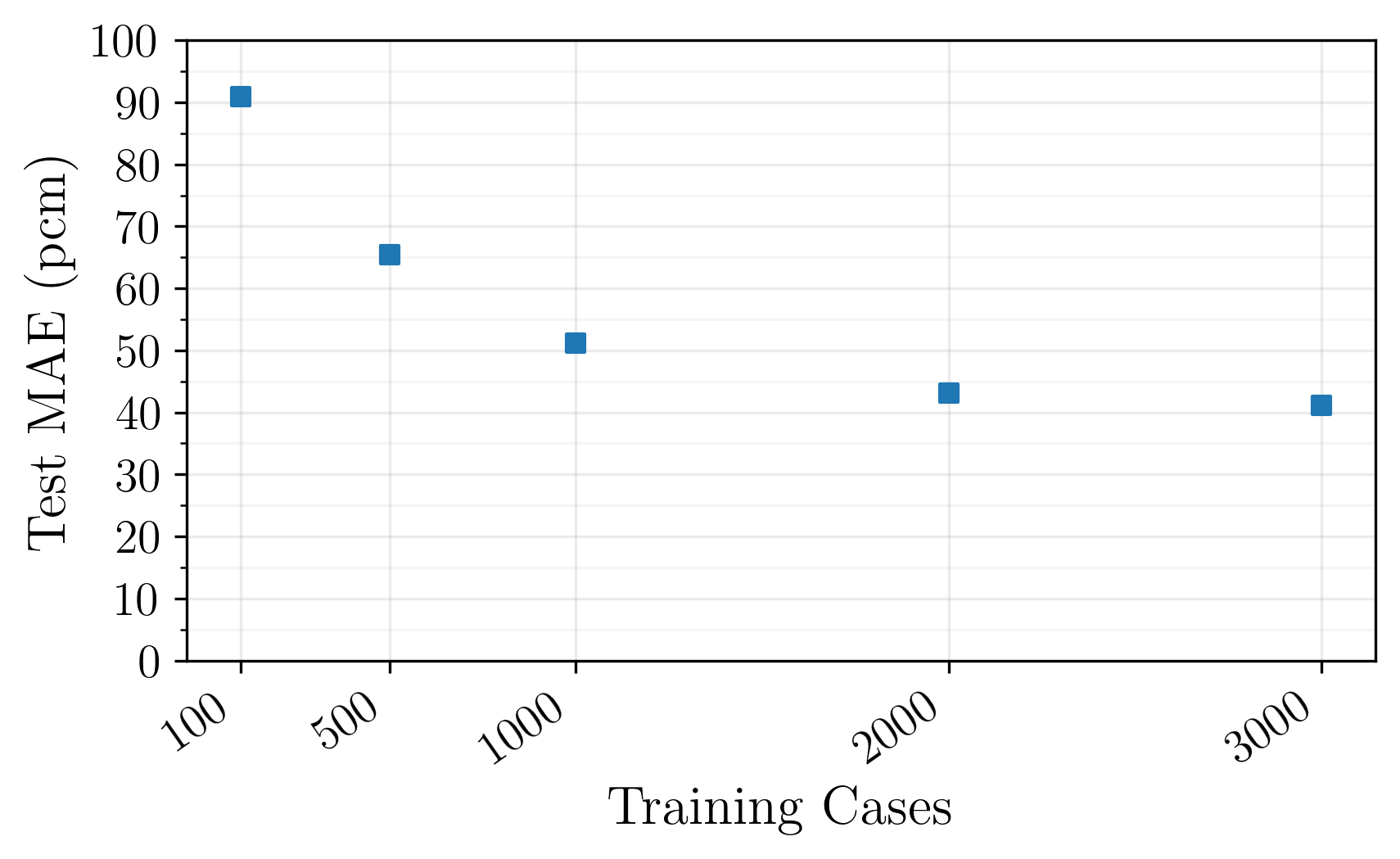}
    \caption{Test set sensitivity error as a function of training set size.}
    \label{fig:ntrain}
\end{figure}
\begin{figure}[H]
    \centering
    \includegraphics[width=0.9\linewidth]{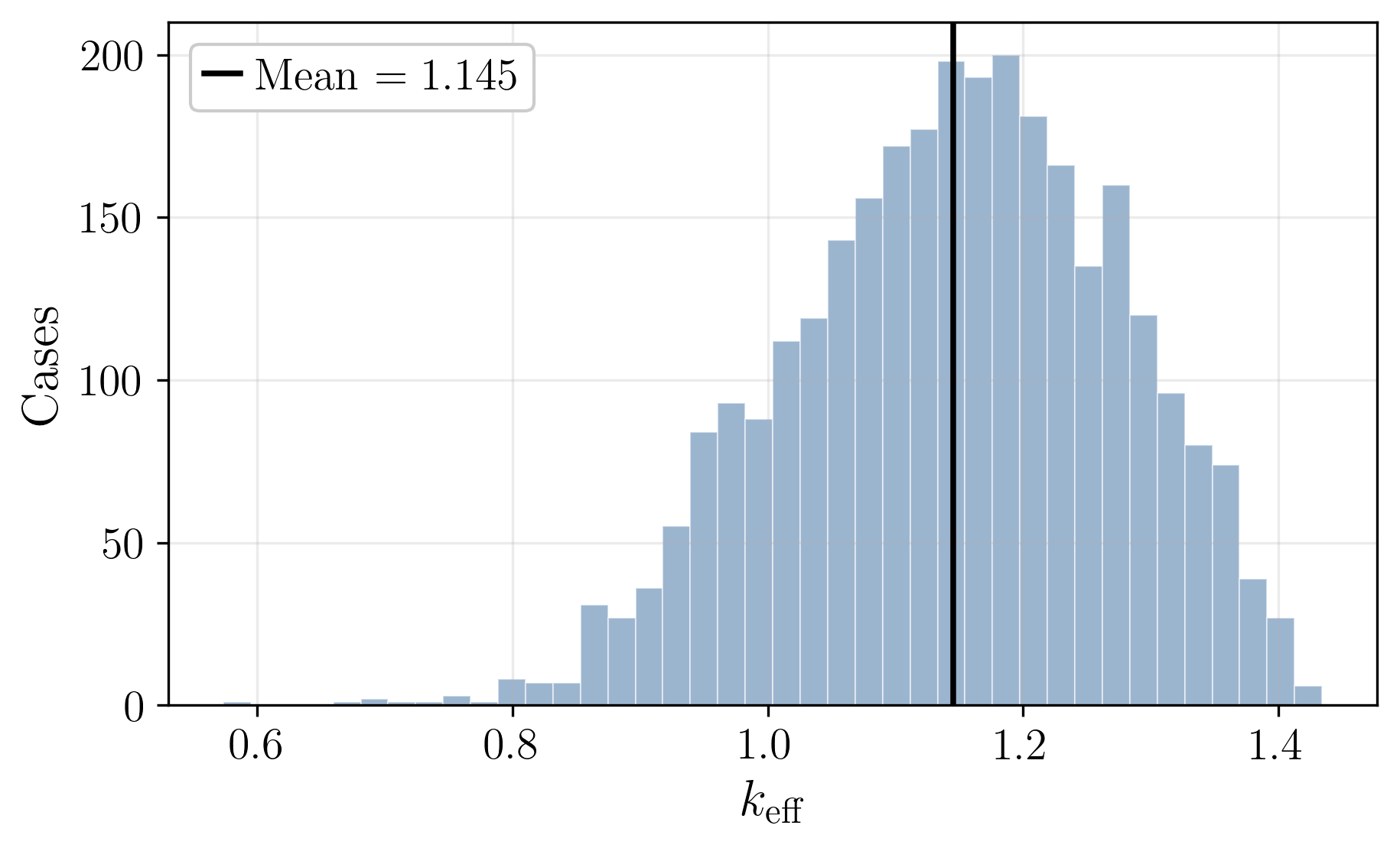}
    \caption{Distribution of $k_\text{eff}$ across the 3,000 training geometries.}
    \label{fig:dataset_keff}
\end{figure}

\begin{figure}[H]
    \centering
    \includegraphics[width=0.9\linewidth]{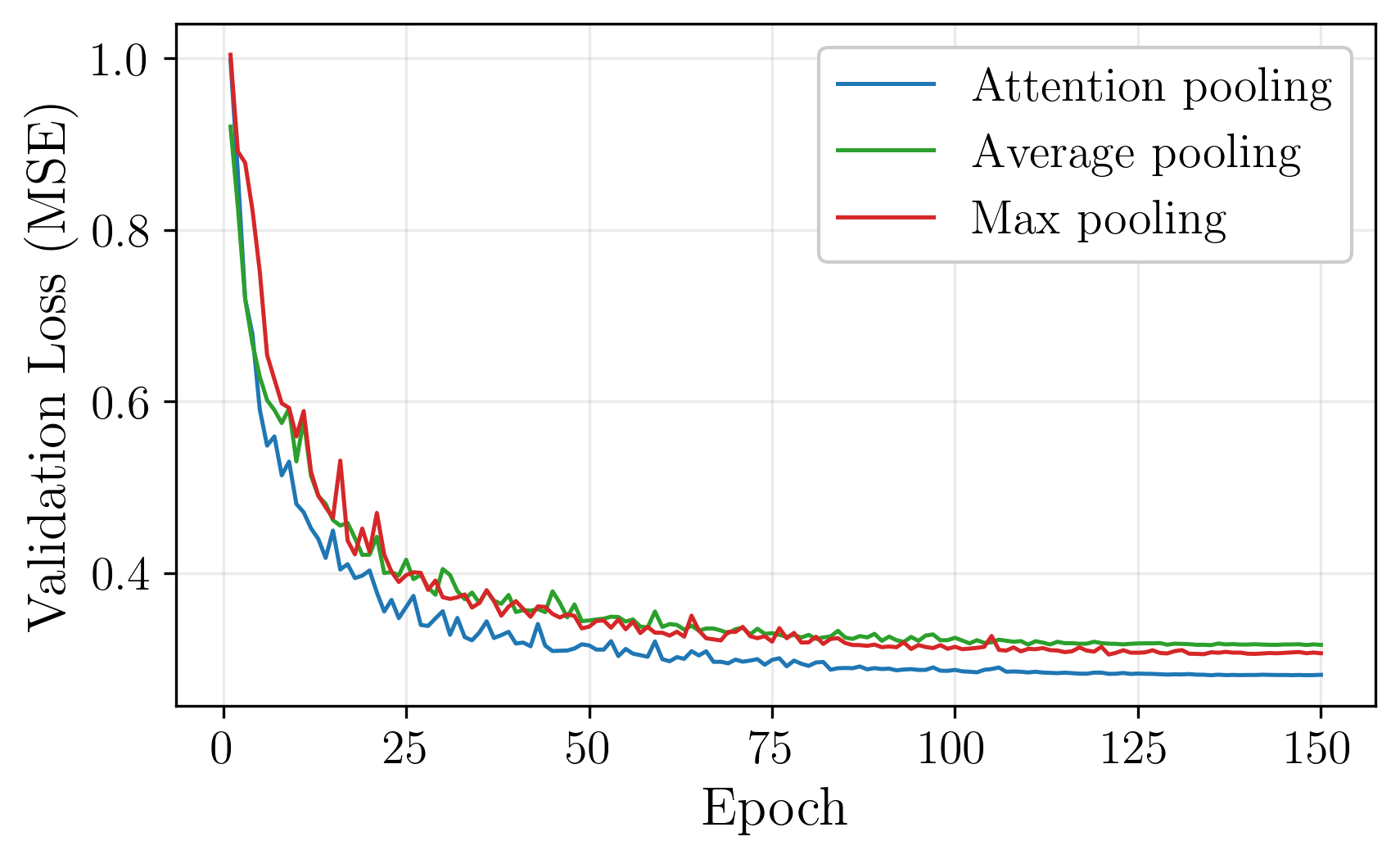}
    \caption{Validation loss curves for different pooling configurations.}
    \label{fig:loss_curves}
\end{figure}

\begin{figure}[H]
    \centering
    \includegraphics[width=0.9\linewidth]{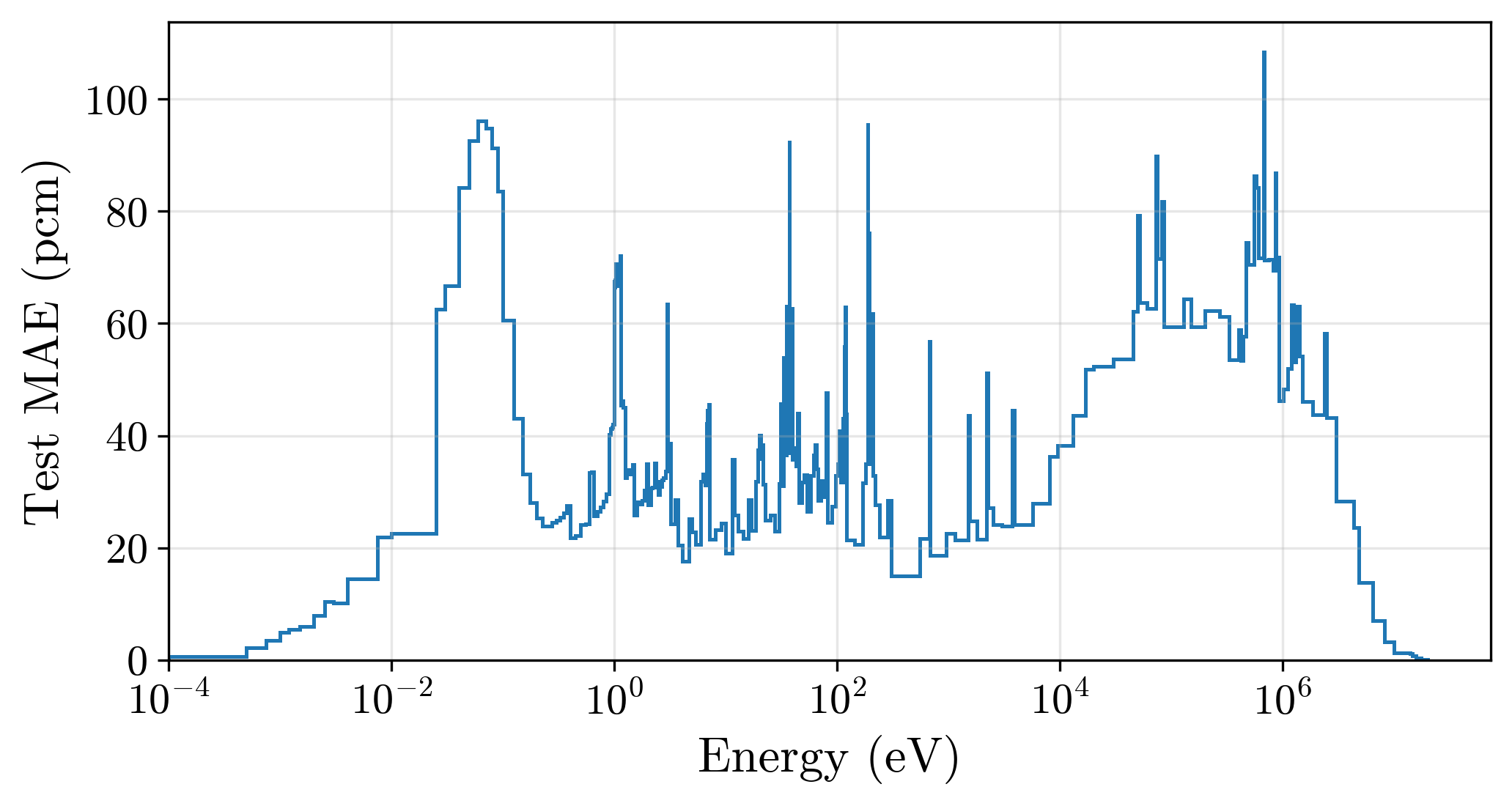}
    \caption{Average test set sensitivity error across all reactions as a function of energy.}
    \label{fig:test_error}
\end{figure}

\begin{figure*}[tb]
    \centering
    \includegraphics[width=1\textwidth]{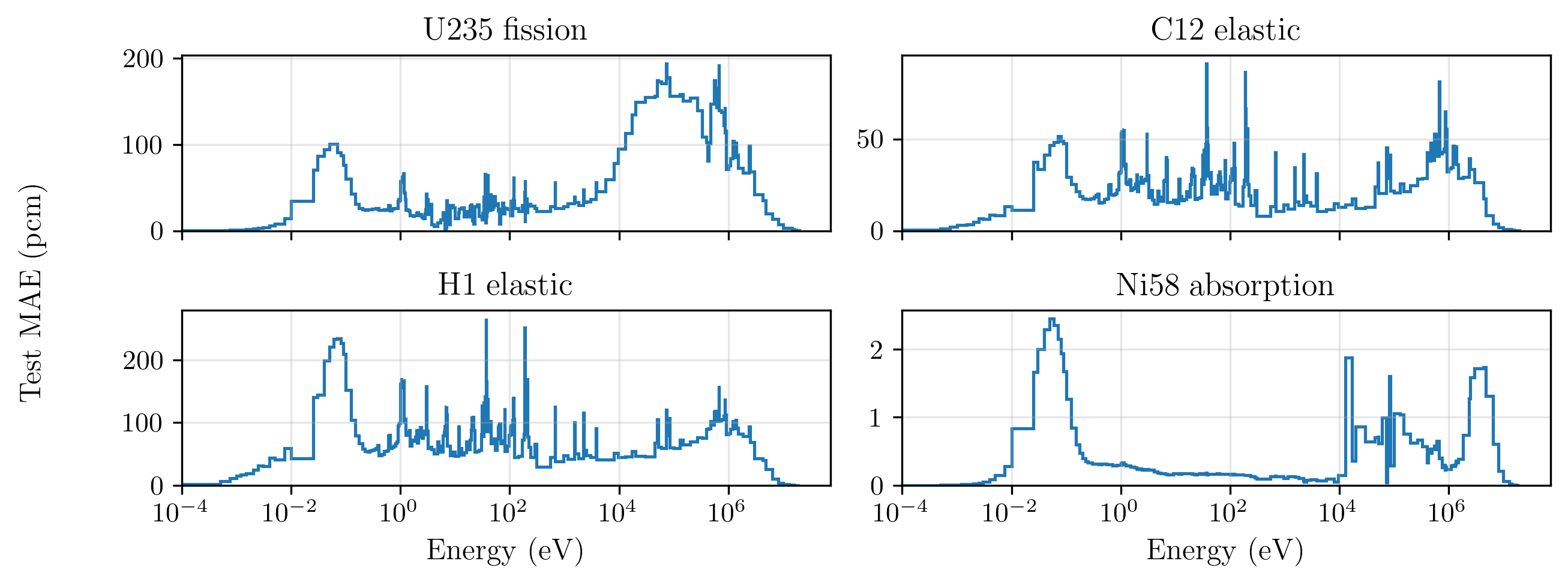}
    \caption{Surrogate model test error (MAE) as a function of reaction and energy group.}
    \label{fig:rxn_error}
\end{figure*}

To benchmark the performance of the attention pooling layer against traditional alternatives, versions of the model were trained and evaluated with max and average pooling.
Each model configuration was evaluated across the training, validation, and test sets.
As shown in Table~\ref{tab:ablation} and Figure~\ref{fig:loss_curves}, attention pooling resulted in the lowest error which supports its further use in the remainder of this work.
Final surrogate error metrics are shown in Table~\ref{tab:val}.
Average test error as a function of energy group is shown in Figure~\ref{fig:test_error}, with per-reaction error shown in Figure~\ref{fig:rxn_error}.
While surrogate error may have some impact on the optimization process, the practice of evaluating multiple of the highest $\hat c_k$ experiment designs in OpenMC ensures that the final chosen geometry is that with the highest true $c_k$ of its peers, despite any errors in $\hat c_k$ prediction.
\begin{table}[H]
\centering
\setlength{\tabcolsep}{3pt}
\renewcommand{\arraystretch}{1.15}
\begin{tabular}{lccc}
\hline
\textbf{Pooling} & \textbf{Train MAE} & \textbf{Val MAE } & \textbf{Test MAE} \\
\hline
Average & 44.63 & 49.48 & 48.27 \\
Max & 44.35 & 47.70 & 46.13 \\
Attention & $\bf 40.01$ & $\bf 42.73$ & $\bf 41.19$ \\
\hline
\end{tabular}
\caption{Sensitivity MAE (pcm) for different pooling configurations with lowest values presented in bold.}
\label{tab:ablation}
\end{table}
\begin{table}[H]
\centering
\setlength{\tabcolsep}{3pt}
\renewcommand{\arraystretch}{1.15}
\begin{tabular}{lccc}
\hline
\textbf{Split} & \textbf{Items} & \textbf{MSE} & \textbf{MAE (pcm)} \\
\hline
Train & 3,000 & $7.076 \cdot 10^{-7}$ & $40.01$ \\
Validation & 1,000& $ 8.051 \cdot 10^{-7}$ & $42.73$ \\
Test & 1,000 & $7.620 \cdot 10^{-7}$ & $41.19$ \\
\hline
\end{tabular}
\caption{Final surrogate error metrics across training, validation, and test sets.}
\label{tab:val}
\end{table}
\begingroup{}
\subsection{Surrogate Performance on Structured Geometries}
\label{subsec:structured}
While very diverse, the geometries generated by the cellular automaton do not cover the entire spectrum of possible geometries.
In particular, the dataset does not cover very ordered and regular geometries, and the test set error metrics above are evaluated only on geometries generated by the cellular automaton.
As discussed above, the surrogate can only be evaluated on geometries that fit the fixed grid, size, and material palette, so experiments from standard references such as the ICSBEP handbook cannot be used as inputs.
To examine performance of the surrogate model on more ordered inputs that do fit these constraints, a set of 100 structured geometries was created, featuring shapes such as concentric squares, circles, checkerboards, and triangles.
Examples of items from the structured set are shown in Figure~\ref{fig:benchmark}.
When evaluated on this set, the surrogate had an average MSE of $1.634\cdot 10^{-5}$ and an average MAE of $145.1\,$pcm, which is approximately 3.5 times the test set MAE of $41.19\,$pcm.
As shown in Figure~\ref{fig:benchmark_error}, most structured geometries have an MAE below $200\,$pcm, but the surrogate performed poorly on a few highly out-of-distribution geometries.
The structured geometries with the ten lowest and ten highest errors are shown in~\ref{sec:app_benchmark}.
The surrogate seems to struggle the most on very finely heterogeneous lattices, while exhibiting the lowest error on geometries comprised of large, heterogeneous regions.
Since ICSBEP experiments cannot be represented on the fixed grid, this structured set is the only test of the surrogate on geometries that were not generated by the cellular automaton.
A more robust approach to future dataset creation would be to combine many approaches and algorithms for geometry generation to further increase coverage of the design space, which may improve generalization to out-of-distribution geometries.
\begin{figure*}[tb]
    \centering
    \includegraphics[width=0.8\textwidth]{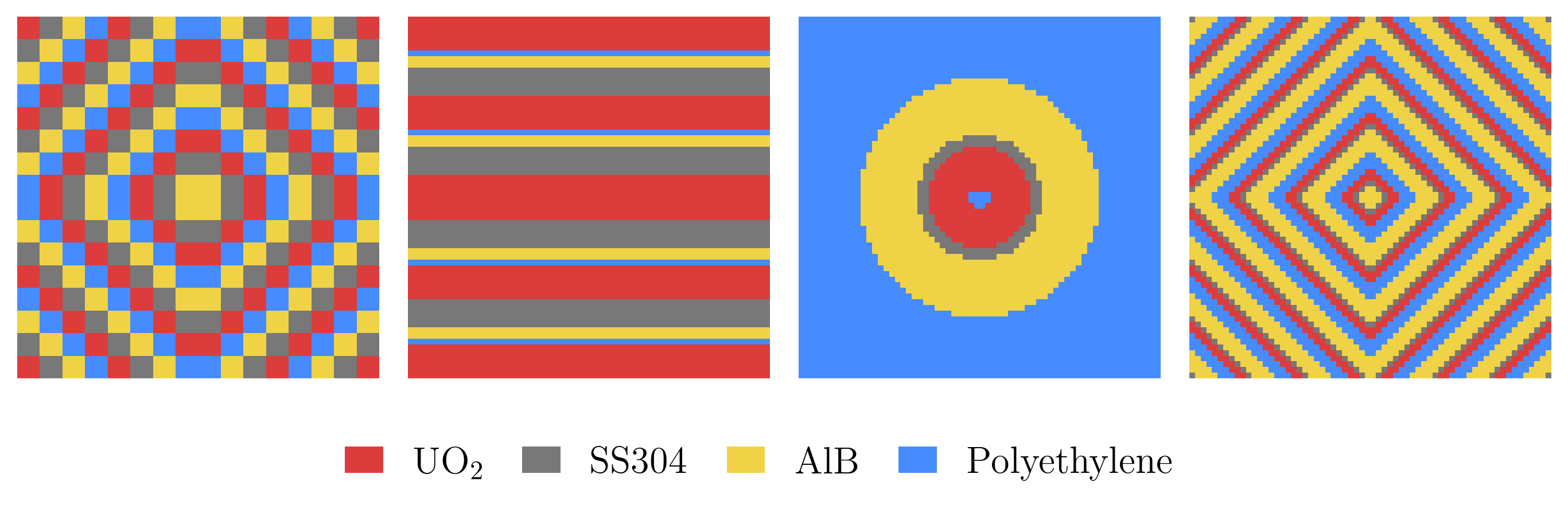}
    \caption{Selected geometries from the set of 100 highly structured geometries.}
    \label{fig:benchmark}
\end{figure*}
\begin{figure}[H]
    \centering
    \includegraphics[width=0.9\linewidth]{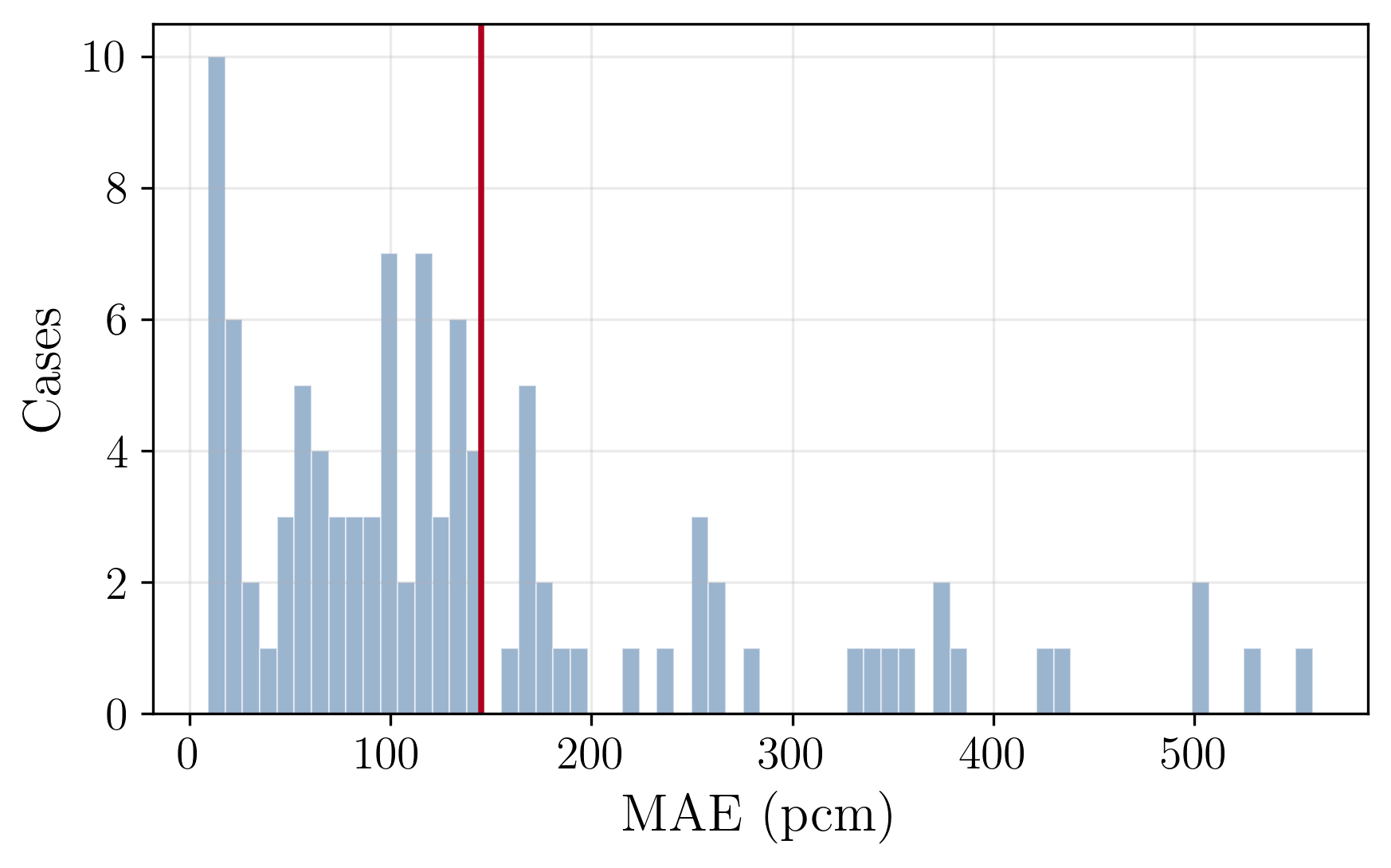}
    \caption{Distribution of surrogate model test error (MAE) evaluated on 100 highly structured geometries. The vertical line marks the average MAE of $145.1\,$pcm.}
    \label{fig:benchmark_error}
\end{figure}
\endgroup{}

\subsection{Interpretability of Multigroup Attention Pools}
\label{subsec:interp}
\begin{figure}[H]
    \centering
    \includegraphics[width=1\linewidth]{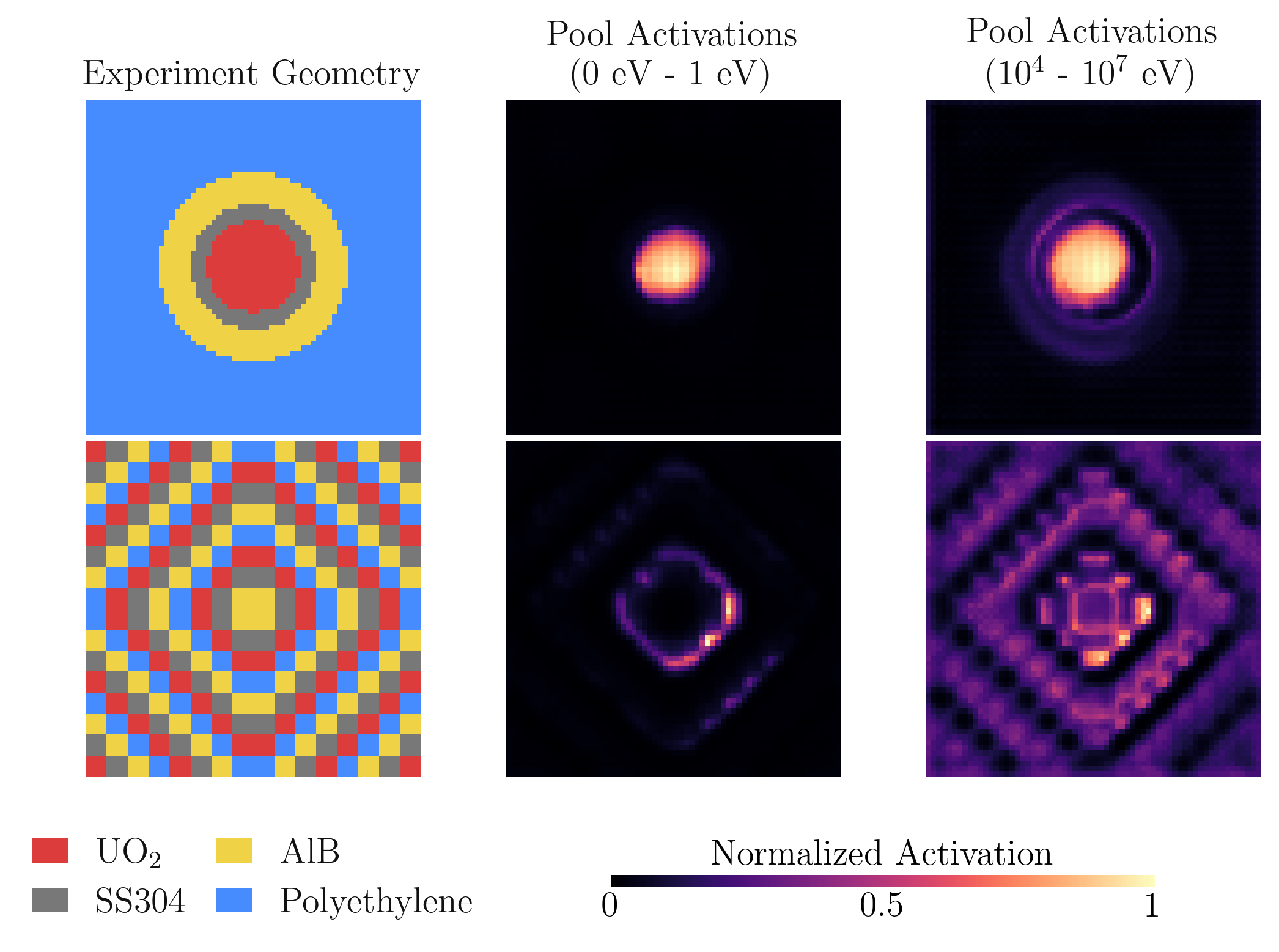}
    \caption{Normalized attention pool activations for $^{235}$U fission sensitivities.}
    \label{fig:attn}
\end{figure}

The internal workings of deep learning models are generally considered ``black box,'' and difficult (or impossible) to interpret.
However, the spatial and physically motivated structure of the multigroup attention pooling layer introduced in Section~\ref{sec:map} means that it can be examined with a degree of human intuition. 
To understand the behavior of the attention pooling layer, the model was run on several simple handcrafted geometries shown in Figure~\ref{fig:attn}. 

For each geometry, the attention pool activations for the $^{235}$U fission sensitivities were extracted and averaged for thermal ($0\,$eV to $1\,$eV) and fast ($10^4\,$eV to $10^7\,$eV) groups.
Since the attention scores do not have physically meaningful units, they were not flux-weighted when rebinning.
While flux-weighting would be necessary if converting energies or sensitivities to a coarser group structure, this interpretability study is nonphysical and qualitative in nature.
The goal is simply to gain a directional understanding of how the model internally represents the information it learns.
A few observations can be made from the activation maps shown in Figure~\ref{fig:attn}.
In both geometries, importance to the lower energy groups is almost entirely contained within the fuel regions themselves, while the higher energy group assigns more importance to the areas surrounding the fuel. 
This is consistent with physical intuition: higher energy neutrons will travel further from the source before interacting, while thermal neutrons are more likely to stay contained within their source.

Additionally, the low energy activations of the second geometry assign more importance to the internal fuel regions than those closer to the edges. 
This is also consistent with intuition, as neutrons born in higher leakage regions are less likely to contribute to fission.

While qualitative, these results indicate the potential of multigroup attention pooling as an interpretable neural network component for capturing the underlying physics of neutron transport.
Better understanding the internal behavior of surrogate models for neutron transport is of great interest, and is a target of future work.

\subsection{TN-LC Experiment Design}
\label{subsec:design}
For each target configuration of the TN-LC, a batch of 500 randomly initialized grids of material assignments were created with an average initial $c_k$ shown in Table~\ref{tab:expr_comparison}.
The gradient optimization algorithm (described in Section~\ref{sec:gbo}) was then run for 2000 steps using the AdamW optimizer~\cite{adamw}.

\begin{figure}[H]
    \centering
    \includegraphics[width=1\linewidth]{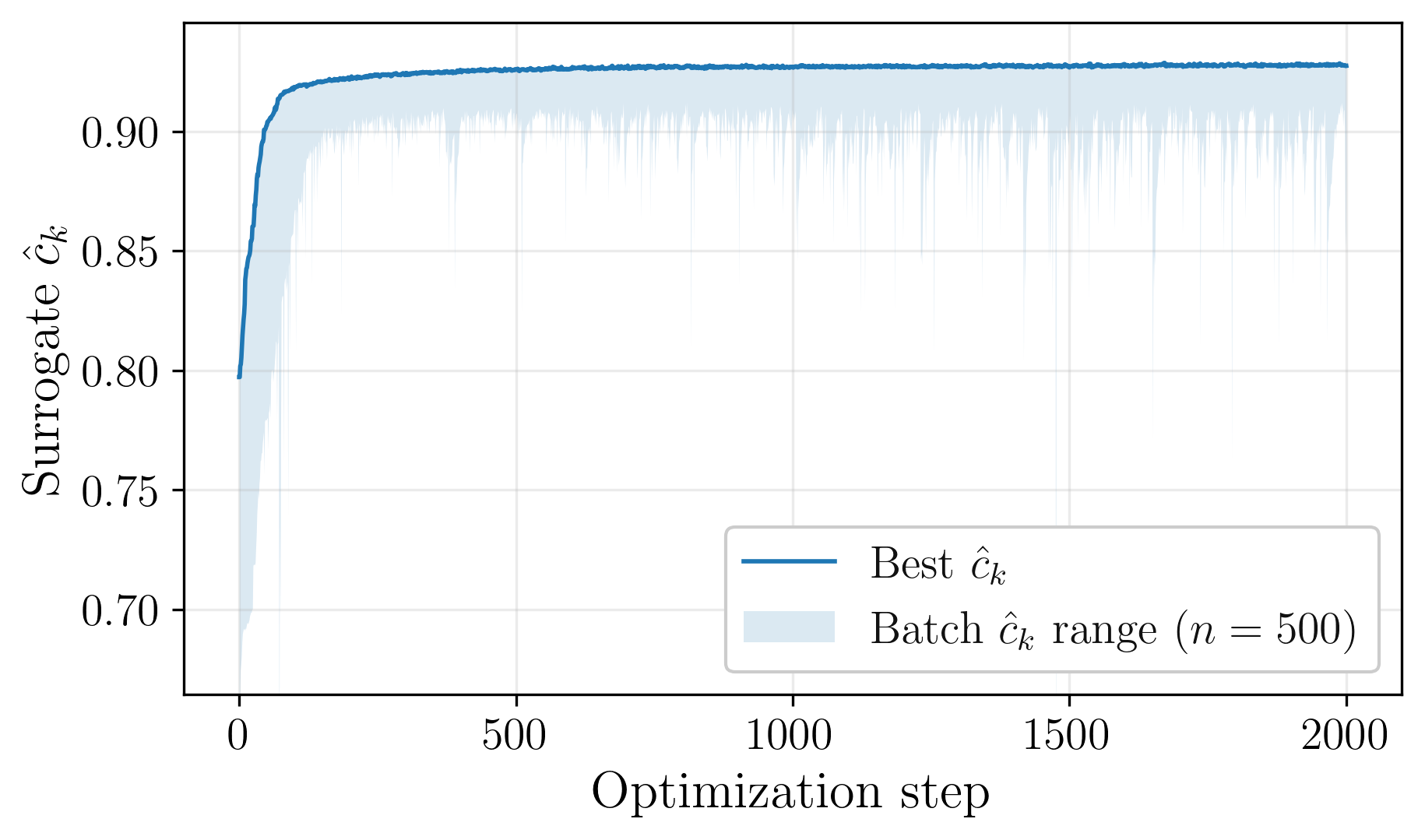}
    \caption{Mean and max $\hat c_k$ at each optimization step during the batch optimization for the Case 2 target configuration.}
    \label{fig:ck_t}
\end{figure}

The batch range and highest $\hat c_k$ as a function of optimization timestep for the Case 2 configuration are shown in Figure~\ref{fig:ck_t}. 
The magnitude of difference between the mean and highest $\hat c_k$ within the batch of 500 initializations indicates the value of optimizing over a large set of geometries, rather than relying on a single initialization.
\begin{figure}[H]
    \centering
    \includegraphics[width=1\linewidth]{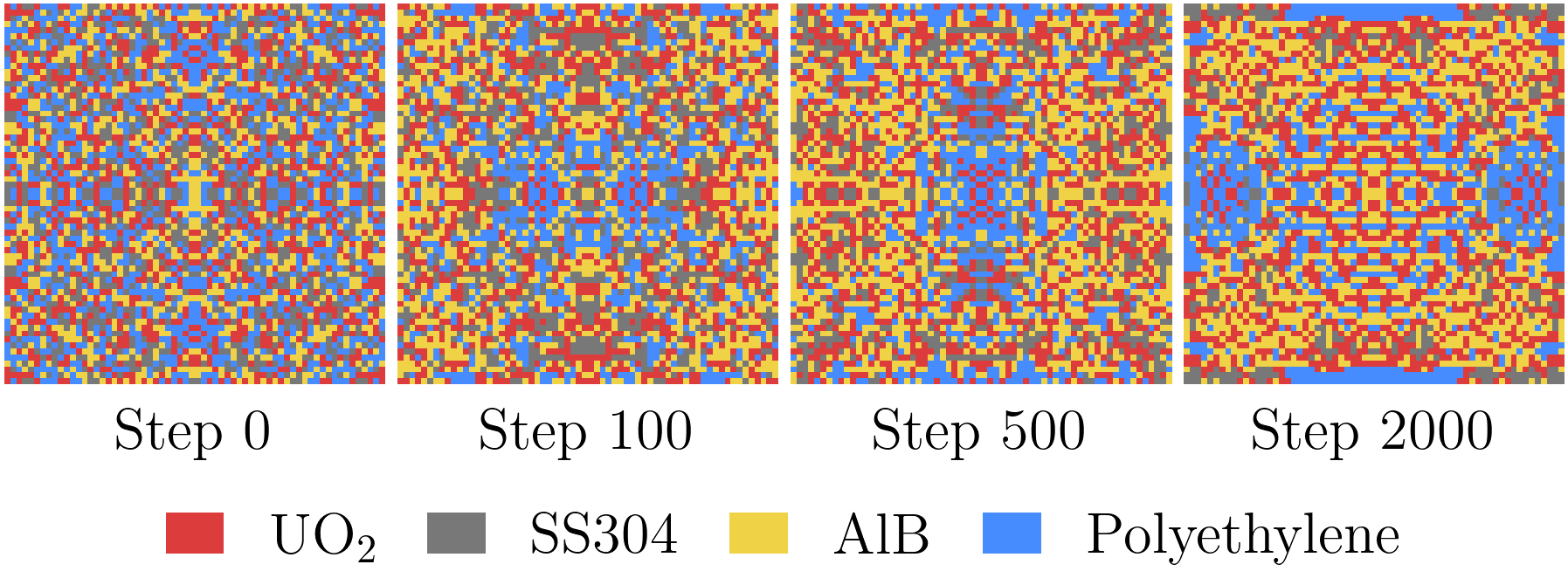}
    \caption{Snapshots of the geometry with the highest $\hat c_k$ at various optimization steps for the Case 2 target configuration.}
    \label{fig:pck_snapshots}
\end{figure}

\begin{figure}[H]
    \centering
    \includegraphics[width=1\linewidth]{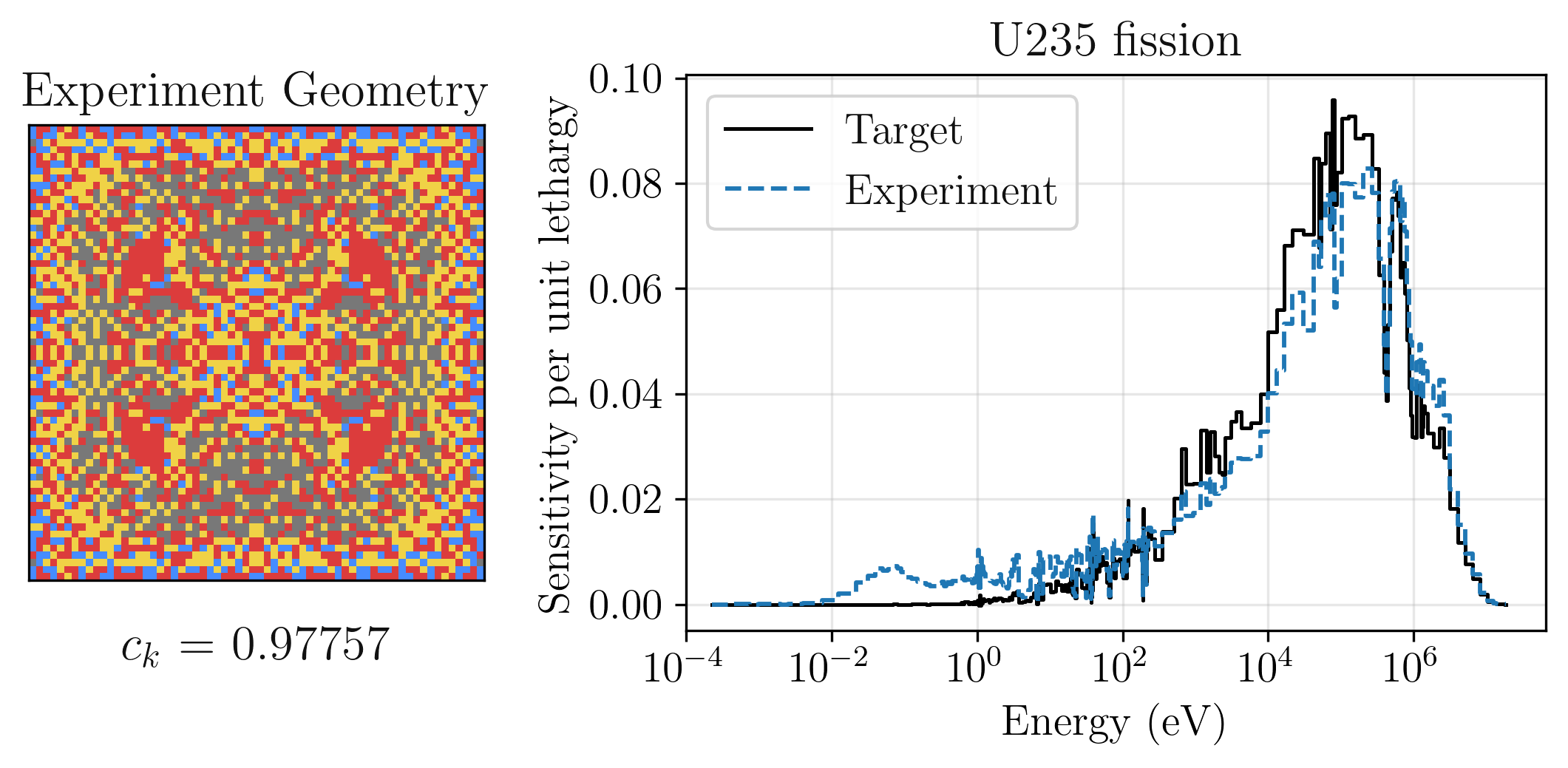}
    \caption{Optimized experiment geometry for the dry TN-LC configuration.}
    \label{fig:tnlc_void}
\end{figure}

\begin{figure}[H]
    \centering
    \includegraphics[width=1\linewidth]{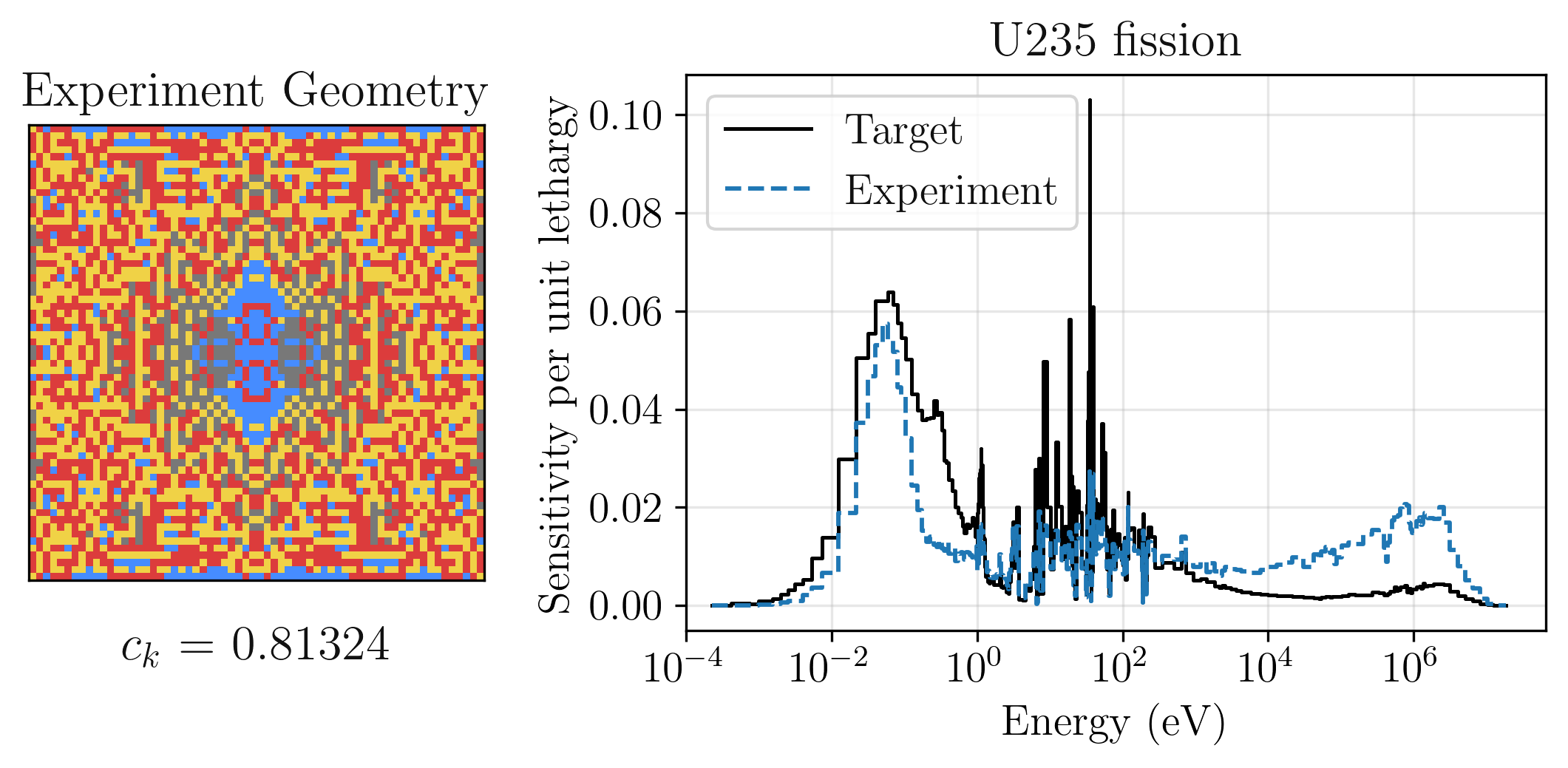}
    \caption{Optimized experiment geometry for the TN-LC Case 1 configuration (low $^{10}\text{B}$ areal density).}
    \label{fig:tnlc_case1}
\end{figure}

\begin{figure}[H]
    \centering
    \includegraphics[width=1\linewidth]{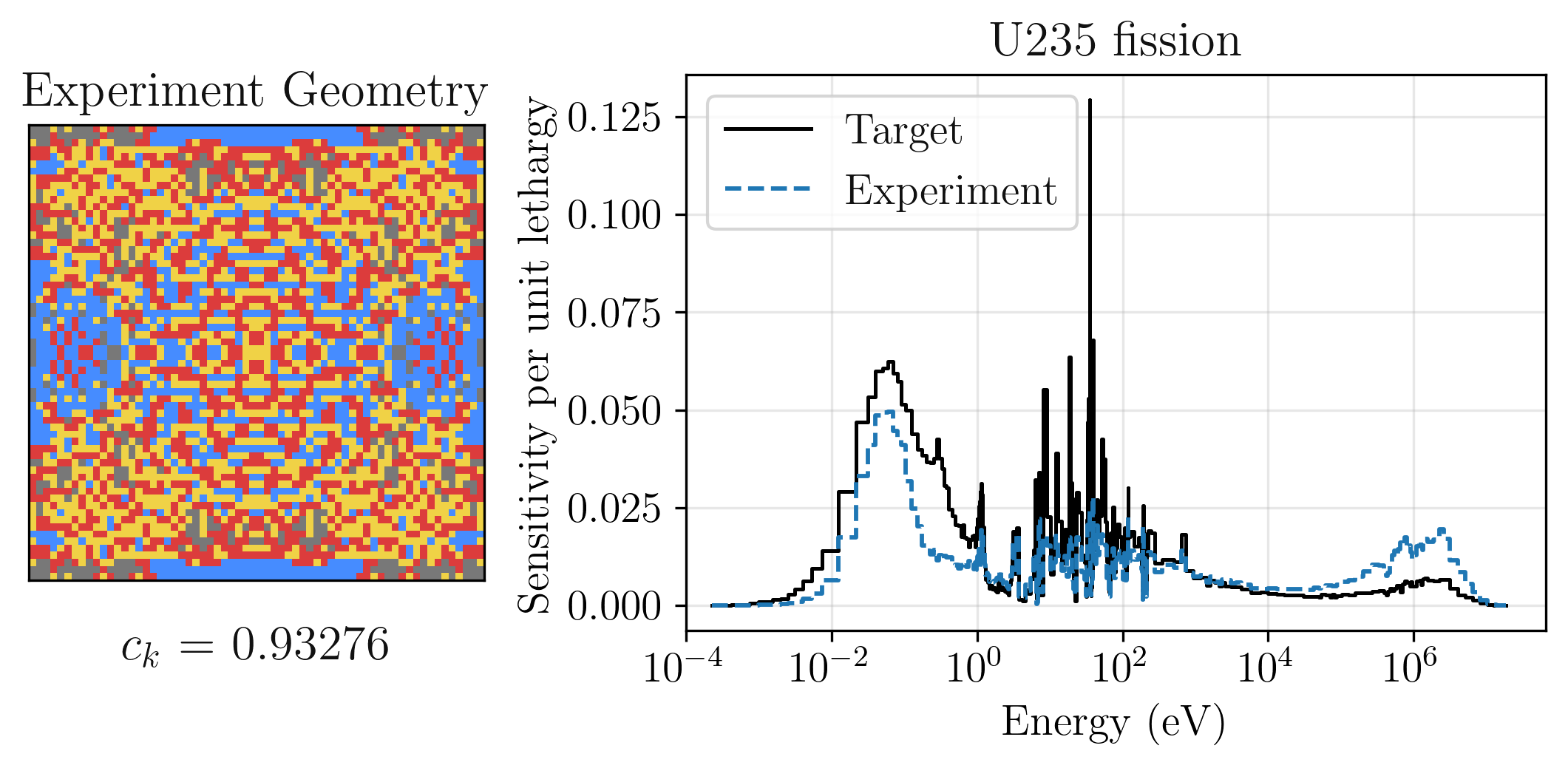}
    \caption{Optimized experiment geometry for the TN-LC Case 2 configuration (high $^{10}\text{B}$ areal density).}
    \label{fig:tnlc_case2}
\end{figure}

Optimized experiment geometries and corresponding $^{235}$U sensitivity vectors are shown for the Dry Cask, Case 1, and Case 2 TN-LC configurations in Figures~\ref{fig:tnlc_void},~\ref{fig:tnlc_case1}, and~\ref{fig:tnlc_case2} respectively, with full sensitivity profiles for all four reactions given in~\ref{sec:app_profiles}.
\begin{table}[H]
\centering
\setlength{\tabcolsep}{3pt}
\renewcommand{\arraystretch}{1.15}
\begin{tabular}{lccc}
\hline
 & \textbf{Dry Cask} & \textbf{Case 1} & \textbf{Case 2} \\
\hline
Mean Initial $\hat c_k$ & $0.86083$ & $0.67963$ & $0.70278$ \\
Final $\hat c_k$ (Predicted) & $0.98227$ & $0.84560$ & $0.92693$ \\
Final $c_k$ (OpenMC) & $0.97757$ & $0.81324$ & $0.93276$ \\
Optimization $c_k$ Gain & 0.11674 & 0.13361 & 0.22998 \\
Experiment $k_\text{eff}$ & 0.97152 & 1.06533 & 1.07930  \\
\hline
\end{tabular}
\caption{Optimization results for the Dry Cask, Case 1, and Case 2 TN-LC configurations.}
\label{tab:expr_comparison}
\end{table}

The optimization procedure was successful in all three cases, yielding experiments with $c_k \geq 0.9$ for the Dry Cask and Case 2 TN-LC configurations and $c_k >0.8$ for the Case 1 configuration.
Notably, a $c_k$ of 0.93276 was achieved for the Case 2 configuration, for which no previous experiment was found with a $c_k \geq 0.8$~\cite{eidelpes2019}.

There are several notable features present in the final geometries.
In each case, the spatial arrangement of materials is more finely heterogeneous than that of a typical training geometry.
Additionally, relatively little stainless steel or polyethylene are present in the final geometries; the optimizer seems to prefer to construct the bulk of the geometries from $\text{UO}_2$ and $\text{AlB}$, as shown in Table~\ref{tab:mats}.
\begin{table}[H]
  \centering
  \setlength{\tabcolsep}{3pt}
  \renewcommand{\arraystretch}{1.15}
  \begin{tabular}{lccc}
  \hline
   & \textbf{Dry Cask} & \textbf{Case 1} & \textbf{Case 2} \\
  \hline
  UO$_2$ & 35.74\% & 38.28\% & 32.03\% \\
  SS304 & 22.66\% & 12.60\% & 9.57\% \\
  AlB & 30.96\% & 38.48\% & 33.30\% \\
  Polyethylene & 10.64\% & 10.64\% & 25.10\% \\
  \hline
  \end{tabular}
  \caption{Material composition percentages in the final optimized geometries
  for the Dry Cask, Case 1, and Case 2 TN-LC configurations.}
  \label{tab:mats}
\end{table}
It is unclear why the Case 1 configuration posed such difficulty for the optimizer, especially given the similarity in sensitivities to Case 2.
Nonetheless, the optimizer successfully increased $c_k$ by 0.13361 to achieve $c_k >0.8$, which Eidelpes et al. set as the cutoff below which a $k_\text{eff}$ penalty must be applied when determining subcritical margins~\cite{eidelpes2019}.

\begin{figure*}[tb]
    \centering
    \includegraphics[width=\textwidth]{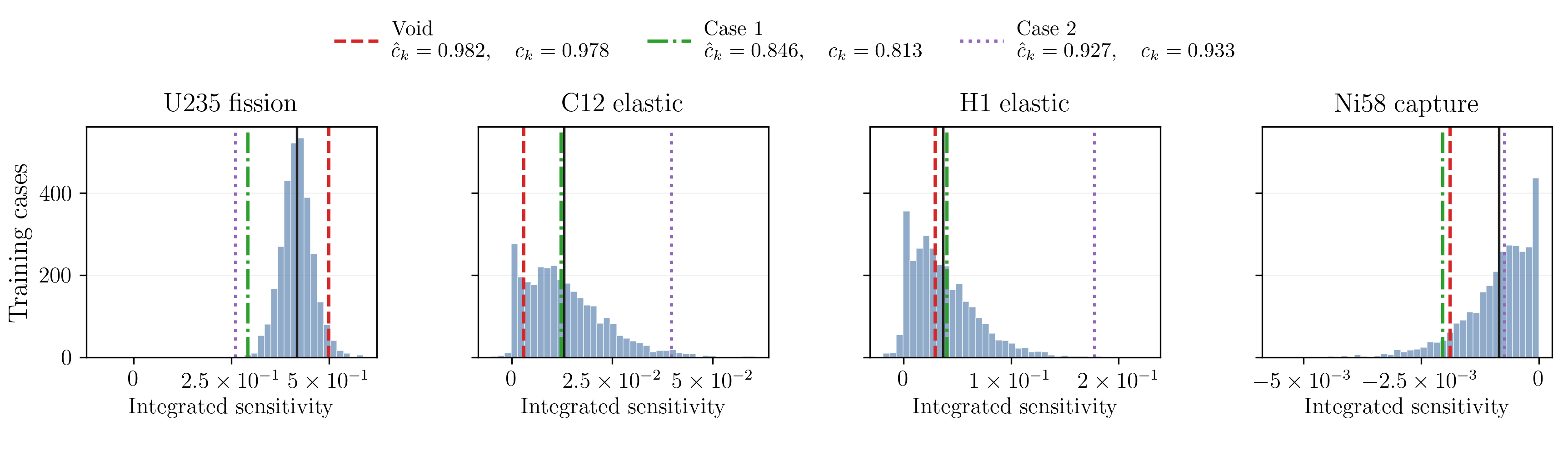}
    \caption{Integrated $k_\text{eff}$ sensitivities of the optimized Dry Cask (Void), Case 1, and Case 2 designs, shown as vertical lines on the distributions and means of the 3,000 training geometries. The predicted $\hat c_k$ and OpenMC-calculated $c_k$ of each design are given in the legend.}
    \label{fig:int_sens}
\end{figure*}

\begin{figure}[H]
    \centering
    \includegraphics[width=1\linewidth]{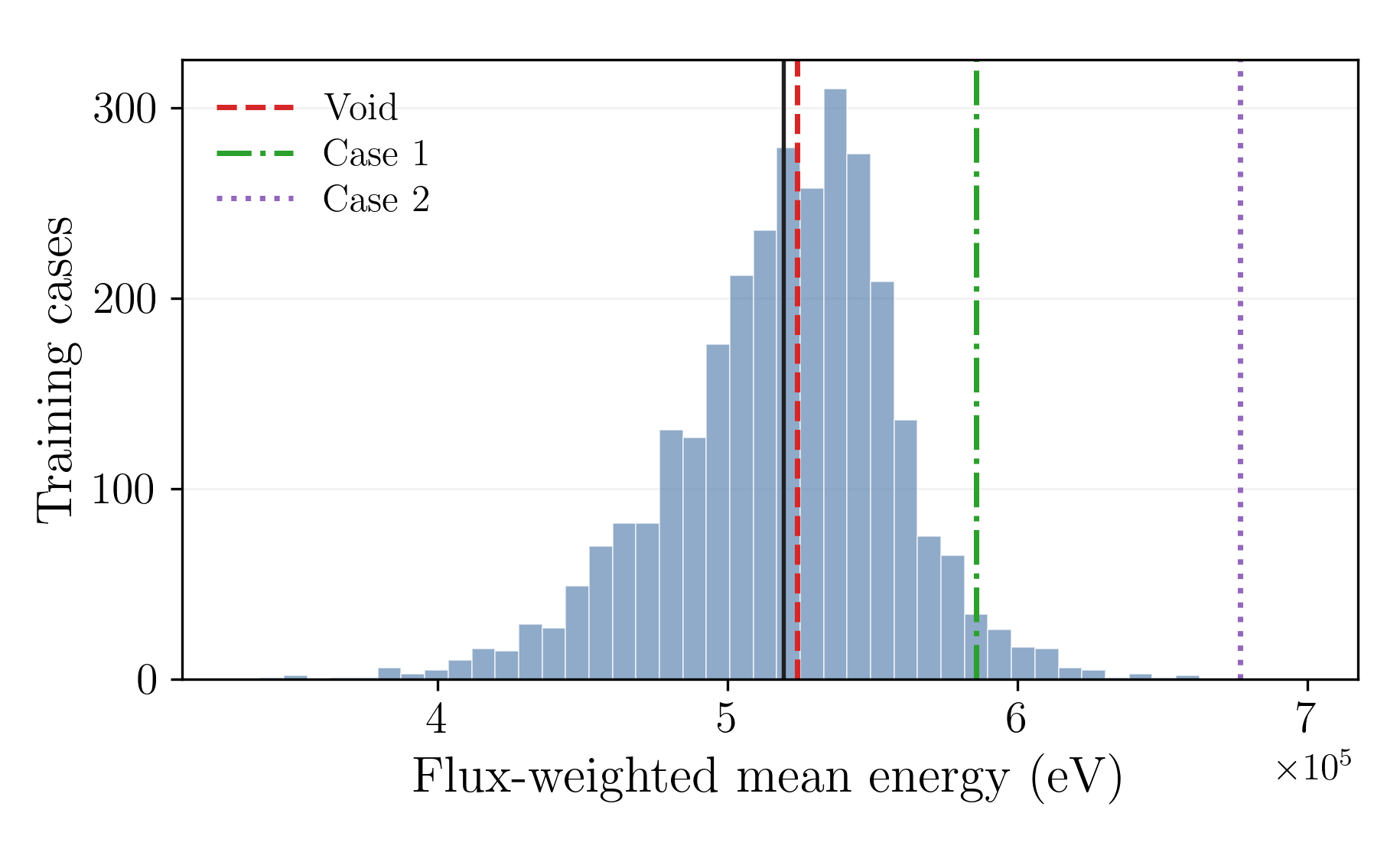}
    \caption{Flux-weighted mean neutron energy $\bar E$ of the optimized Dry Cask (Void), Case 1, and Case 2 designs, shown as vertical lines on the distribution and mean of the 3,000 training geometries.}
    \label{fig:flux_energy}
\end{figure}

\begingroup{}
The position of each optimized design within the training distributions is shown in Figures~\ref{fig:int_sens} and~\ref{fig:flux_energy}.
In contrast to the application cases, all integrated sensitivities of the optimized designs lie within the range of the training data, as does the $\bar E$ of the Dry Cask and Case 1 designs.
The Case 2 design has a harder spectrum than any training geometry, with $\bar E$ of 676.8\,keV compared to a training maximum of 662.6\,keV.
Several design values also lie in the tails of the training distributions, such as the integrated $^{235}$U fission sensitivities of the Case 1 and Case 2 designs and the integrated $^{1}$H elastic sensitivity of the Case 2 design.
The optimization increased $c_k$ in all three cases, while the designs remained largely within the neutronic regime of the training data.

Because $c_k$ is normalized by the magnitudes of both sensitivity profiles in Equation~\ref{eq:ckdefinition}, it is unchanged by a uniform scaling of either profile.
In addition, the $^{235}$U fission sensitivity is the primary contributor to $k_\text{eff}$ uncertainty, and is therefore expected to contribute most to $c_k$.
As a result, a design can achieve a high $c_k$ while its integrated sensitivities, particularly those of the other reactions, differ from those of the application case.
Nonetheless, the optimized designs follow several trends of the application cases across the three configurations.
The integrated $^{1}$H elastic sensitivities of both the application cases and the designs increase from the Dry Cask to Case 1 to Case 2.
The Dry Cask application case has the most negative integrated $^{58}$Ni capture sensitivity, and the Dry Cask design contains the most stainless steel, as shown in Table~\ref{tab:mats}.
Similarly, the Case 2 application case has the least negative integrated $^{58}$Ni capture sensitivity, and the Case 2 design contains the least stainless steel.
The Case 1 design, which has the lowest $c_k$, contains the same polyethylene fraction as the Dry Cask design and has a similar integrated $^{1}$H elastic sensitivity, while the Case 2 design contains more than twice as much polyethylene.
Although both flooded application cases are strongly thermal, only the Case 2 design is substantially more moderated than the Dry Cask design, as measured by its polyethylene fraction and integrated $^{1}$H elastic sensitivity.
This may partly explain the difficulty of the Case 1 optimization.
The mean initial $\hat c_k$ was also lower for the two flooded configurations than for the Dry Cask, as shown in Table~\ref{tab:expr_comparison}, indicating that randomly initialized geometries are predicted to be less similar to the flooded configurations.
\endgroup{}


The degree to which the different reactions and energies contribute to $c_k$, especially given the effect of covariance weighting, is not immediately visible when visually comparing sensitivity vectors.
Depending on the sensitivity of particular energy groups, bias in shape can have little effect on $c_k$ when those groups contribute minimally to the relevant sensitivity or covariance-weighted uncertainty.
Overlaid sensitivity curves should therefore be treated as qualitative evaluations of similarity, with the understanding that only $c_k$ can provide a quantitative assessment.


To explore the full capabilities of the method unencumbered by human biases, no constraints were imposed on the design space other than the material palette and grid dimensions.
As such, the optimized geometries are unintuitive and intentionally ``alien'' in design.
The optimization algorithm considers only the single-objective problem of maximizing $\hat c_k$ without regard for factors such as cost, construction feasibility, leakage, or $k_\text{eff}$ itself.
However, these additional constraints can be enforced by the introduction of additional terms to the loss function of the optimizer.
In particular, optimization of $k_\text{eff}$ would require training a separate surrogate model, which would be straightforward but is beyond the focus of this work.
As stated in Section~\ref{sec:intro}, these results are a proof-of-concept of an unconstrained optimization method.

\section{Conclusion}
\label{sec:conc}
This work presents a proof-of-concept study of the inverse design of nuclear criticality experiments using deep neural network surrogate modeling and gradient-based geometry optimization.
This study evaluates the preliminary performance of an unconstrained optimization method within a fixed design space.
A dataset was created from 5,000 procedurally generated experiment geometries and corresponding sensitivity profiles, which was then used to train a surrogate model for sensitivity prediction.
The model combined a U-Net~\cite{unet} with a novel multigroup attention pooling layer, introduced to capture the importance of different regions in the geometry to the sensitivity of each energy group.
The attention pooling layer demonstrated superior predictive performance relative to standard pooling alternatives, as well as internal representations qualitatively consistent with the intuitively expected neutron flux distribution across each tested geometry.

A gradient optimization algorithm was used to design criticality experiments given a target sensitivity profile with the maximization of the correlation coefficient $c_k$ as the optimization objective.
Applied to the design of experiments for the validation of the TN-LC HALEU transportation cask, the optimization procedure successfully produced experiment geometries with $c_k$ greater than 0.9 for two of three considered configurations, notably including a flooded cask scenario for which no previous experiment was identified with $c_k>0.8$~\cite{eidelpes2019}.
The correlation coefficients for the dry cask, flooded cask with low-density poison, and flooded cask with high-density poison were 0.97757, 0.81324, and 0.93276 respectively.
The experiment $k_\text{eff}$ values were 0.97152, 1.06533, and 1.07930.

\begingroup{}
The training dataset was characterized using integrated reaction sensitivities, which describe the fuel, moderation, and absorption conditions of each geometry.
The flux-weighted mean energy was also used as a spectral indicator.
The flooded application cases have a larger thermal fraction of $^{235}$U fission sensitivity than any training geometry, and the application cases lie outside the training data for several descriptors.
The optimized designs lie within the training range of each descriptor, except for the flux-weighted mean energy of the Case 2 design, which exceeds the training maximum.
Additionally, the surrogate MAE increased from $41.19\,$pcm to $145.1\,$pcm when evaluated on 100 structured geometries.
Surrogate accuracy on ICSBEP experiments could not be measured, since these experiments cannot be represented on the fixed grid.
Every reported $c_k$ is calculated in OpenMC, so the reported correlation coefficients do not depend on the accuracy of the surrogate.
\endgroup{}

The primary shortcoming of the method is that it does not consider $k_\text{eff}$ during the optimization process.
The maximization of $c_k$ is the sole optimization objective, without constraints on $k_\text{eff}$, construction feasibility, or material cost. 
For realistic implementation, $k_\text{eff}$ must be considered as part of the optimization process to ensure that all generated experiments reach $k_\text{eff}\geq1$.
The incorporation of these considerations through additional surrogate models or terms in the optimization objective is a natural and tractable extension of the method.

\begingroup{}
A methodology for practical critical experiment design would require several further developments.
The training data should be expanded to include structured geometries, grid-based approximations of existing benchmark experiments, and strongly thermal geometries.
The surrogate should also provide an estimate of its own uncertainty, for example through model ensembles or Gaussian process models~\cite{gp}.
This would allow the optimizer to be penalized for leaving the region supported by the training data.
\endgroup{}
Extension to three-dimensional geometries, variable experiment size, and the full material design space of a critical experiment facility such as SPARC~\cite{woolstenhulme2025} are also directions for future work.
With these developments, the approach could aid experiment designers by proposing candidate configurations with high $c_k$, which could then be refined and verified using established methods.
Additionally, more rigorous interpretability analysis could be done to quantify the behavior of multigroup attention pooling.

The ability to rapidly propose candidate critical experiment designs with high $c_k$ for arbitrary target systems has the potential to accelerate the validation of advanced nuclear technology for which historical experimental coverage is limited.
The performance and observed internal behavior of the surrogate model in this work indicates the potential of deep learning applied to neutronics.
There is potential for further work to explore the degree to which high-fidelity surrogates can be trained to approximate more fundamental properties of neutron transport.
This would allow the approach presented in this work to be applied to a wider range of nuclear system design tasks.

\section*{Acknowledgement}
This research made use of the resources of the High Performance Computing Center at Idaho National Laboratory, which is supported by the Office of Nuclear Energy of the U.S. Department of Energy and the Nuclear Science User Facilities under Contract No. DE-AC07-05ID14517.


\bibliographystyle{elsarticle-num} 
\bibliography{refs.bib}

\appendix

\section{Surrogate Architecture and Hyperparameters}
\label{sec:app_hparams}
\begin{table}[h]
\centering
\caption{Surrogate architecture and training hyperparameters.}
\label{tab:hparams}
\begin{tabular}{lll}
\hline
\textbf{Parameter} & \textbf{Value} \\
\hline
\multicolumn{3}{l}{\textit{Input}} \\
Geometry dimensions & $64 \times 64$  \\
One-hot materials & $4$ \\
\hline
\multicolumn{3}{l}{\textit{U-Net}} \\
Channel widths & 32, 64, 128, 256 \\
Conv2d settings & $k=3,\,\, p=1$ \\
GroupNorm settings & $n_g=8$ \\
Dropout & 0.2  \\
\hline
\multicolumn{3}{l}{\textit{Multigroup Attention Pooling}} \\
Spatial dimensions & $64\times64$  \\
Energy group structure & SCALE-252 \\
\hline
\multicolumn{3}{l}{\textit{Group Regressor}} \\
Hidden channels & 32 \\
Residual blocks & 6 \\
Conv1d settings & $k=3,\,\, p=1$ \\
GroupNorm settings & $n_g=8$ \\
\hline
\multicolumn{3}{l}{\textit{Training}} \\
Optimizer & AdamW & \\
Learning rate & $10^{-3} \to 10^{-5}$ \\
Epochs & $150$ & \\
Batch size & $64$ & \\
Loss function & MSE \\
Training items & 3000 \\
Validation items & 1000 \\
Test items & 1000 \\
\hline
\end{tabular}
\end{table}

\section{Selected Structured Benchmark Geometries}
\label{sec:app_benchmark}
\begin{figure}
    \centering
    \includegraphics[width=0.8\textwidth]{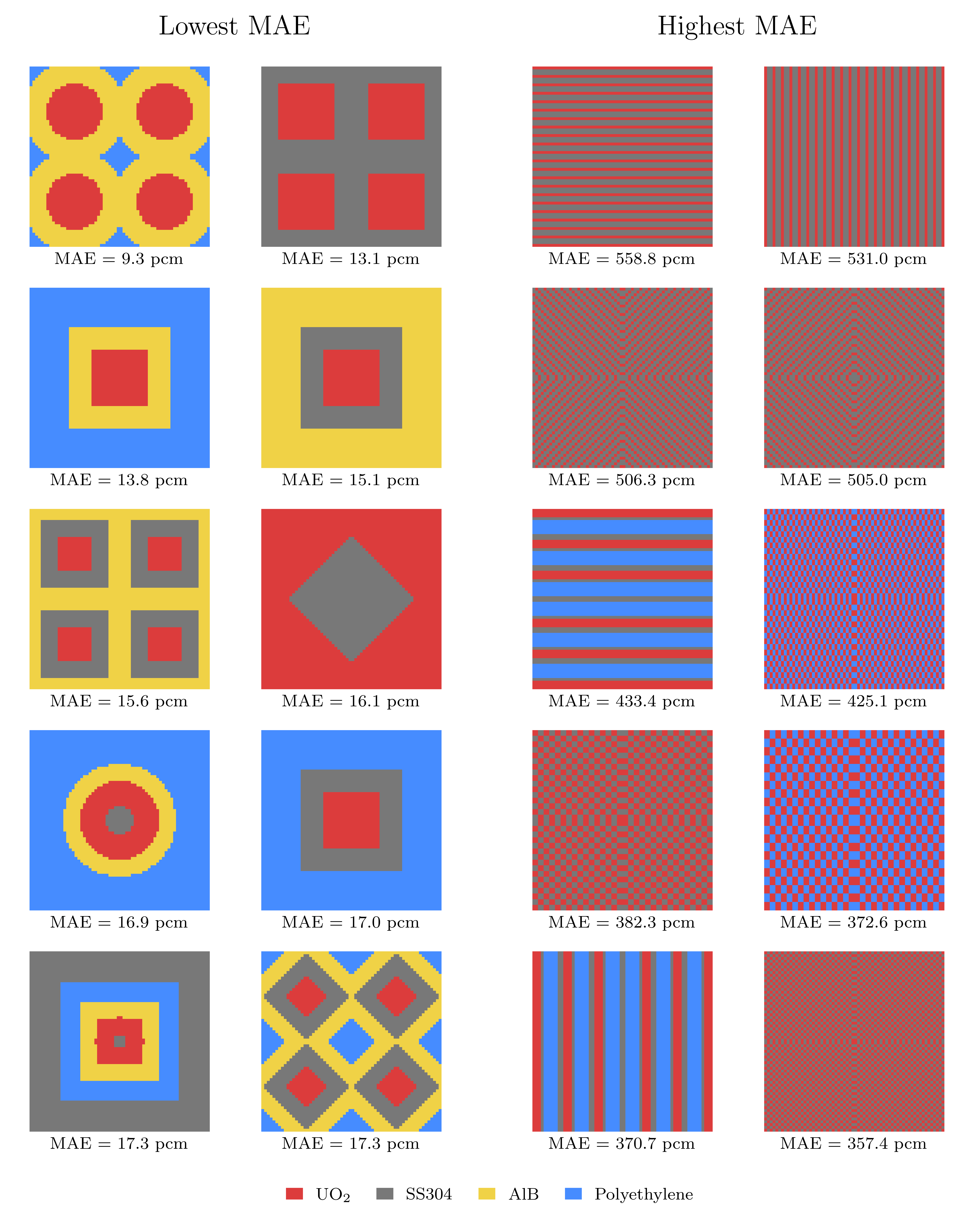}
    \caption{Structured benchmark geometries with the lowest and highest surrogate evaluation error.}
    \label{fig:extremes}
\end{figure}

\section{Sensitivity Profiles For Proposed TN-LC Experiments}
\label{sec:app_profiles}
\begin{figure*}
    \centering
    \includegraphics[width=0.95\textwidth]{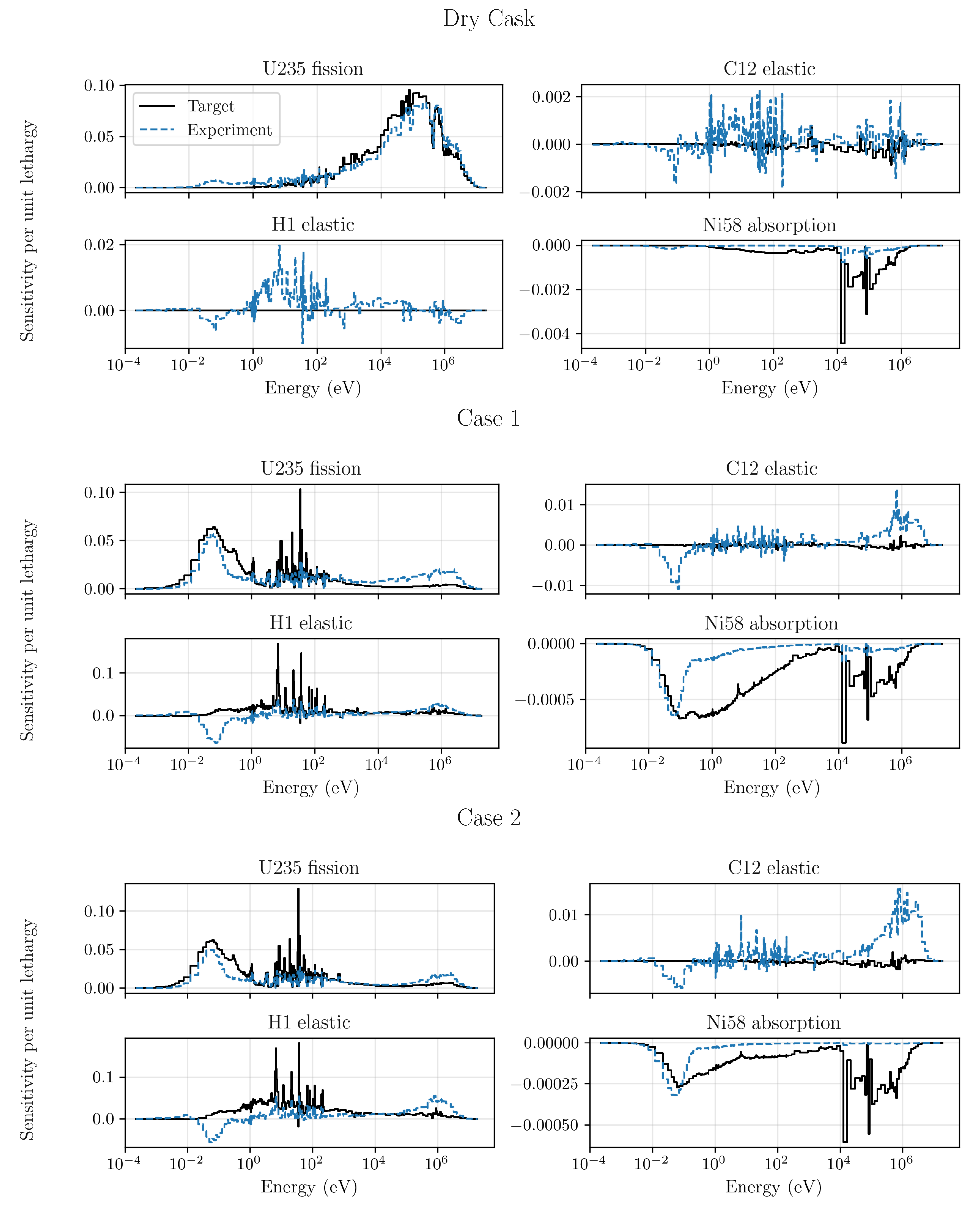}
    \caption{Full sensitivity profiles for the three optimized experiment geometries.}
    \label{fig:all_sens}
\end{figure*}

\end{document}